\documentclass{article} 
\usepackage{iclr2020_conference,times}

\usepackage{amsmath,amsfonts,bm}

\def\eqref#1{equation~\ref{#1}}

\def\1{\bm{1}}

\DeclareMathAlphabet{\mathsfit}{\encodingdefault}{\sfdefault}{m}{sl}
\SetMathAlphabet{\mathsfit}{bold}{\encodingdefault}{\sfdefault}{bx}{n}

\usepackage{hyperref}
\usepackage{url}
 \usepackage{array,multirow,graphicx}
 
\usepackage[utf8]{inputenc} 
\usepackage[T1]{fontenc}    
\usepackage{hyperref}       
\usepackage{url}            
\usepackage{booktabs}       
\usepackage{amsfonts}       
\usepackage{nicefrac}       
\usepackage{microtype}      
\usepackage[noend, boxruled]{algorithm2e}
\usepackage{amsmath}
\usepackage{bbm}
\usepackage{graphicx}
\usepackage{caption}
\usepackage{subcaption}
\usepackage{makecell}
\usepackage{tabularx}
\usepackage{float}
\usepackage{xr}
\iclrfinalcopy

\newcolumntype{+}{>{\global\let\currentrowstyle\relax}}
\newcolumntype{^}{>{\currentrowstyle}}

\newcolumntype{L}{>{\raggedright\arraybackslash}X}
\usepackage[usenames, dvipsnames]{color}
\newcommand{\innercolorunderline}[4]{%
  \vtop{
    \offinterlineskip
    \sbox0{$#4$}
    \copy0
    \kern#1
    \hbox{\color{#3}\vrule height 1pt width \wd0}
    \kern#2
  }%
}

\ExplSyntaxOn
\keys_define:nn { diane/colorunderline }
 {
  color .tl_set:N  = \l_diane_colorunderline_color_tl,
  color .initial:n = red,
  above .dim_set:N = \l_diane_colorunderline_above_dim,
  above .initial:n = 1pt,
  below .dim_set:N = \l_diane_colorunderline_below_dim,
  below .initial:n = 0pt,
 }

\NewDocumentCommand{\colorunderline}{O{}m}
 {
  \group_begin:
  \keys_set:nn { diane/colorunderline } { #1 }
  \diane_colorunderline:VVVn
    \l_diane_colorunderline_above_dim
    \l_diane_colorunderline_below_dim
    \l_diane_colorunderline_color_tl
    { #2 }
  \group_end:
 }
\cs_set_eq:NN \diane_colorunderline:nnnn \innercolorunderline
\cs_generate_variant:Nn \diane_colorunderline:nnnn { VVV }
\ExplSyntaxOff

\usepackage{algorithm2e}
\title{EDUCE: Explaining model Decisions through Unsupervised Concepts Extraction}

\author{Diane Bouchacourt \\
  Facebook AI Research\\
  \texttt{dianeb@fb.com} \\
  \And
  Ludovic Denoyer \\
    Facebook AI Research\\
    \texttt{denoyer@fb.com} \\
}
\usepackage{xcolor}

\begin{document}
\maketitle
\begin{abstract}
Providing explanations along with predictions is crucial in some text processing tasks. Therefore, we propose a new self-interpretable model that performs output prediction and simultaneously provides an explanation in terms of the presence of particular concepts in the input. To do so, our model's prediction relies solely on a low-dimensional binary representation of the input, where each feature denotes the presence or absence of concepts. The presence of a concept is decided from an excerpt i.e. a small sequence of consecutive words in the text. Relevant concepts for the prediction task at hand are automatically defined by our model, avoiding the need for concept-level annotations. To ease interpretability, we enforce that for each concept, the corresponding excerpts share similar semantics and are differentiable from each others. We experimentally demonstrate the relevance of our approach on text classification and multi-sentiment analysis tasks.
\end{abstract}

\section{Introduction}

While deep learning models are powerful tools to perform a large variety of tasks, their predictive process often remains obscure. Understanding their behavior becomes crucial. This is particularly true with text data, where predicting without justifications has limited applicability. There has been a recent focus on trying to make deep models more interpretable, see for example \citep{Ribeiro:etal:2016, Bach:etal:2015,Shrikumar:etal:2017,Simonyan:etal:2014,Sundararajan:etal:2017}. Specifically, methods have been recently proposed to provide such explanations \emph{simultaneously} with the prediction. For example, \citep{Lei:etal:2016,Yu:etal:2019,Bastings:etal:2019} select subsets of words in the input text that can account for the model's prediction (called rationales). In this work, we propose a model that provides an explanation based on the absence or presence ``concepts'' that are automatically discovered in the texts.

\begin{figure}[t!]
\begin{center}
\includegraphics[width=\textwidth]{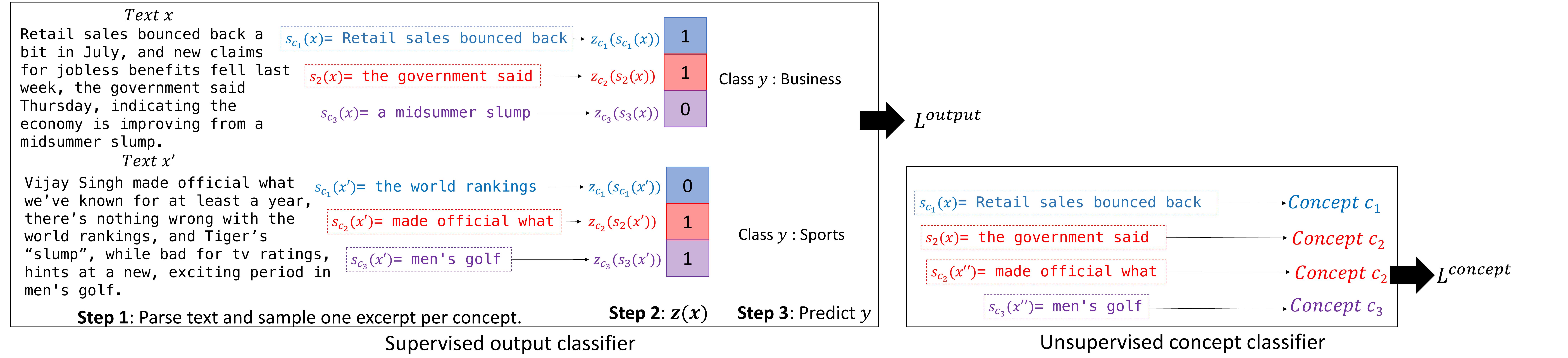}
\vspace{-7mm}
\end{center}
\caption{Illustration of EDUCE's prediction process on AGNews topic classification dataset \citep{Zhang:LeCun:2015} examples. The different steps are explained in the text.}
\vspace{-5mm}
\label{fig:ludo1}
\end{figure}

Suppose we ask a user to tell what is the category (or class) of the top text of Figure \ref{fig:ludo1} (text $x$). She detects that the words
\emph{the government said} relate to a specific concept that is present in $x$, a concept she also detect in the words \emph{made official what} in text $x'$ . Let us call this concept ``politics'' in the remainder. She also notes that the words \emph{retails sales bounced back} refer to a concept different from the previous one. Let us call this other concept ``economy''. As the text is concerned with politics and economy she infers that its category is Business. Said otherwise, she detects excerpts that relate to particular concepts and decides on the text category based on the detected concepts.

Similarly, our paradigm assumes that an explanation of a model's prediction is understandable if it relies on a few concepts, where each concept relates to parts of text (referred to as \emph{excerpts}) that are semantically consistent across multiple texts. Our methodology is as follows. First, our model encodes the input text into a binary low-dimensional vector, where each value (0 or 1) is computed from an excerpt and denotes the presence or absence of a specific concept in the input. Then, the prediction is made from this binary vector alone, allowing an easy interpretation of the decision in term of presence/absence of concepts.

While we set a maximal value of the number of concepts (which is the dimensionality of the binary representation), the concepts are \textbf{unsupervised and not defined a priori}. The model automatically determines them in a way that eases interpretation: each concept is encouraged to be semantically consistent and not to overlap with other concepts. Therefore, extracted excerpts for a concept must be discriminative of that concept only and share similar semantics. This is enforced through a concept consistency and separability constraint added as an auxiliary loss in the learning optimization problem. As a result, each discovered concept can be understood from the corresponding excerpts extracted that activate its appearance: in our previous example, the meaning of the first concept the user identifies is inferred from the excerpts she detected for that concept in $x$ and $x'$, i.e. \emph{the government said, made official what}. Looking at these excerpts, we identify that concept as politics.

Our idea relates to Latent Dirichlet Allocation (LDA) \citep{Blei:etal:2013}, where text documents are described by a set of topics that are semantically consistent. However, LDA builds a probabilistic model of text generation, whereas our goal is to discover and define latent concepts that are relevant for a prediction task at hand. In comparison to rationale-based text processing models \citep{Lei:etal:2016,Yu:etal:2019,Bastings:etal:2019}, we rely on a different paradigm: our model's prediction is based on the absence or presence of discovered concepts, and makes no direct use of the words captured as excerpts. We do so to ease interpretation of the prediction. This makes the interpretation of a different nature than these methods, and simpler to understand for the user.

Our contribution with this work is a new self-interpretable model that predicts solely from the presence/absence of concepts in the input. Concepts are learned unsupervisingly, and described by a semantically consistent set of excerpts. We experiment on three text categorization tasks and a multi-aspect sentiment analysis task, compare our model's performance with state-of-the-art prediction models, and demonstrate its interpretability. Note that an instance of our model for image processing is described in the supplementary
Section \ref{imludo} with illustrative experiments.

\section{The EDUCE model}
\label{sec:model}

We present our model, called EDUCE for Explaining model Decisions through Unsupervised Concepts Extraction, using a multi-class classification task, but our method can be used to perform regression or multi-label classification. We consider a training dataset $\mathcal{D}$ of inputs $(x_1,....,x_N)$, $x_n \in \mathcal{X}$ and corresponding labels $(y_1,...,y_N)$, $y_n \in \mathcal{Y}$.

Contrarily to back-box models that use complex computations over low-level input features to predict the output class $y$, our objective is to enforce the model to map an $x$ input to an easy-to-interpret representation $\bm{z}(x) \in \{0,1\}^C$, on which the output prediction relies. EDUCE's inference process consists in the following steps. \textbf{Step 1}: For each concept $c$ (i.e. each dimension of $\bm{z}(x)$), compute $p_{\gamma}(s|x,c)$: the probability of each excerpt $s$ in $x$ to be selected. Sample a unique excerpt $s_c(x) \sim p_{\gamma}(s|x,c)$ per concept (per dimension of $\bm{z}(x)$). Note that the same excerpt can be selected for multiple concepts. \textbf{Step 2}: Decide on each value $z_c(x)$ of the representation $\bm{z}(x)$ by sampling from $p_{\alpha}(z_c|s_c(x),c)$. Given $s_c(x)$, $z_c(s_c(x))=1$ means concept $c$ is detected as present and $z_c(s_c(x))=0$ means it is absent. \textbf{Step 3}: Predict the output class $y$ from $\bm{z}(x)=(z_1(s_1(x)),...,z_C(s_C(x))$).

We do not have concept-level annotations. To ensure semantic consistency and prevent overlap of the concepts, we jointly train a concept classifier to recognize, for every excerpt $s_c(x)$ such that $z_c(s_c(x))=1$, the concept (i.e. the dimension) it was extracted for: the label for each $s_c(x)$ is simply $c \in [1, C]$. Figure \ref{fig:ludo1} illustrates these steps, which we detail below.

\subsection{Generating predictions and explanations}
\label{sec:modelstep1}

\textbf{Step 1: Extract a unique excerpt $s_c(x)$ per concept $c$.} Given an input sentence $x:w_1, ..., w_M$, where each $w_i$ is a word, a excerpt $s$ consists of a span of consecutive words of flexible size between 3 and 10 words: $s=w_k,...,w_{k+l}, 3\le l\le10$. For each concept, we extract a unique excerpt, defined by its first and last word. Our extraction process is very similar to the one proposed in Question-Answering models (e.g \citet{Devlin:etal:2019}). An excerpt is sampled by first sampling its \textit{start word} and then its \textit{end word} conditioned on the start word.

We denote the probability over each excerpt $s$ to represent concept $c$ in $x$ as $p_{\gamma}(s|x,c)$ and we write as the product of (i) $p_{start}(w_k|x,c)$, the probability for $s$'s first word to be the start word of the excerpt to extract and (ii) $p_{stop}(w_{k+l}|x,c,w_k)$ the probability of $w_{k+l}$ to be the stop word given the start word $w_k$:
$p_{\gamma}(s|x,c)=p_{\gamma}(w_k,...,w_{k+l}|x,c)=p_{stop}(w_{k+l}|x,c,w_k)p_{start}(w_k|x,c)$. We parametrize $p_{start}(w_{k}|x,c)$ and $p_{stop}(w_{k+l}|x,c,w_k)$ using recurrent neural networks. First we feed the entire input sentence $x:w_1, ..., w_M$ through a bidirectional LSTM, and we represent each word $w_i$ in the sentence as the concatenation of the forward pass and backward pass hidden states for that word: $h_k=[\overrightarrow{h_k},\overleftarrow{h_k}],~\forall k \in [1,M]$. Second, each $h_k$ is fed to a linear layer with parameters $\gamma_{start}$ which outputs a score for $w_k$, followed by the softmax activation function over all possible words to have the probability distribution over each word to be the \emph{start word}:
\begin{equation}
p_{start}(w_k|x,c)=\dfrac{\exp(\gamma_{start} \cdot h_k)}{\sum_{k=1}^M\exp(\gamma_{start} \cdot h_k)},~k=1, ..., M.
\end{equation}
Using this distribution, we can sample a specific start word $w_{start} \sim p_{start}(w_k|x,c)$. Third, we feed again the vectors $h_k$ to another linear layer with parameters $\gamma_{stop}$ that gives a score for each word to be the \emph{stop word}. We mask out these scores before taking the softmax, such that the probabilities for $w_{start+l}$ with $l < 3$ and $l > 10$ to be the \emph{stop word} are $0$.
\[
  p_{stop}(w_{start+l}|x,c,w_{start})=\begin{cases}
               \dfrac{\exp(\gamma_{stop} \cdot h_{start+l})}{\sum_{l=3}^{10}\exp(\gamma_{stop} \cdot h_{start+l})},~l\ge 3~\text{or}~l\le10\\
               0, ~l<3~\text{or}~l>10.
            \end{cases}
\]
We then sample a specific stop word $w_{stop} \sim p_{stop}(w_{start+l}|x,c,w_{start})$\footnote{We also prevent stop words to be outside of the text length.}. We now have the start word $w_{start}$ and stop word $w_{stop}$. The corresponding excerpt is $w_{start},...,w_{stop}$ and is represented as a fixed size vector $s_c(x)$ by computing the average of (pre-trained, fixed) words embeddings vectors of $w_{start},...,w_{stop}$. Hence sampled excerpts $s_c(x)$ belong to $\mathbb{R}^d$, where $d$ is the dimension of embedding vectors.

This process is illustrated in Figure \ref{fig:ludo1}: For text $x$, the excerpt \emph{Retails sales bounced back} is selected for concept $c_1$, \emph{the government said} for concept $c_2$ and \emph{a midsummer slump} for concept $c_3$.

\textbf{Step 2: For each $c$, from the excerpt $s_c(x)$, decide on the value $z_c(x)$ denoting the presence/absence of concept $c$.} Each extracted excerpt (one per concept) is processed to decide on the absence or presence of each concept in $x$. Specifically, for each concept $c$ we take the dot product of a weight vector $\alpha_c \in \mathbb{R}^d$ with $s_c(x)$, followed by a sigmoid activation function, in order to obtain the Bernoulli probability
\begin{equation}
p_{\alpha}(z_c=1|s_c(x),c)=\sigma(\alpha_c \cdot s_c(x))
\label{eq:step3}
\end{equation}
This is the probability that $s_c(x)$ extracted from $x$ for concept $c$ triggers the presence of concept $c$. The binary vector $\bm{z}(x)$ is obtained by independently sampling each dimension: $\forall c~z_c \sim p_{\alpha}(z_c|s_c(x),c)$.

This step can be seen as $C$ independent binary classifiers, each of them deciding on the presence of a particular concept. We illustrate this in Step 2 in Figure \ref{fig:ludo1}: For text $x$, the excerpt \emph{Retails sales bounced back} activates the presence of concept $c_1$, \emph{the government said} activates the presence of $c_2$ but \emph{a midsummer slump} does not activate the presence of $c_3$.

\textbf{Step 3: Predict $y$ from $\bm{z}(x)$.} As shown in Step 3 of Figure \ref{fig:ludo1}, given an input $x$, the prediction of the output $y$ is made solely from the intermediate binary representation $\bm{z}(x)$. We use a linear classifier without bias, parametrized by a weight matrix $\delta \in \mathbb{R}^{\mathcal{|Y| \times C}}$ followed by a softmax activation function, returning $p_\delta(y|\bm{z}(x)) \forall y \in \mathcal{Y}$.

\textbf{Output classification training objective.} The parameters $\{\gamma, \alpha, \delta\}$ are learned in a end-to-end manner by minimizing cross-entropy, which writes for each $x$ as:
\begin{flalign}
\mathcal{L}^{output}(x,y, \delta, \alpha, \gamma)&=\mathbb{E}_{\bm{s}(x) \sim p_\gamma}[\mathbb{E}_{\bm{z}(x)\sim p_\alpha}[\mathcal{L}^{output}(y,\bm{z}(x),\delta)]] \nonumber \\
&=\mathbb{E}_{\forall c~s_c(x) \sim p_{\gamma}(s|x,c)}[\mathbb{E}_{\forall c~z_c \sim p_{\alpha}(z_c|s_c(x),c)}[-\log p_\delta(y|\bm{z}(x))]]. \label{eq:ocloss}
\end{flalign}
We give details about the optimization in Section \ref{subsec:obj}.

\subsection{Unsupervised Discovery of Latent Concepts}
\label{subsec:conceptclassif}

To ease interpretation of the concepts' meaning, we want each dimension $\bm{z}$ to correspond to semantically consistent concepts. In other words, for a given concept $c$ (i.e. a given dimension $z_c$), for all inputs where that concept is present (i.e. $\forall x~s.t.~z_c(s_c(x))=1$), the corresponding excerpts $s_{c}(x)$ should share common semantics. While the $C$ binary classifiers deciding on the presence of each concept $c$ (Step 2) enforce consistency within the set of extracted excerpts $s_c$, there is no guarantee that the concepts are not overlapping (i.e. are separable). We want concepts to be separable so that each concept captures a particular notion. Excerpts extracted for concept $c$ must be distinguishable from the excerpts extracted for another concept $c'$. If we take the example of Figure \ref{fig:ludo1}: In Step 2 the excerpt \emph{Retails sales bounced back} is identifiable as relating to economy/finance, contrarily to \emph{the government said, made official what} which trigger the concept politics.

We rephrase our desiderata for consistency and separability as classification of the excerpts, with as many classes as the number of concepts. An external classifier should recognize that elements in the set $\{s_c(x) |\forall x~z_c(s_c(x))=1\}$ belong to concept $c$, and $\{s_{c'}(x) |\forall x~z_{c'}(s_{c'}(x))=1\}$ belong to concept $c'$ and not $c$. Recall that \textbf{we do not have labels for the concepts}. The ground-truth label of each extracted excerpt $s_c(x)$ is the \textbf{index $c$} of the dimension of $\bm{z}(x)$ for which it was extracted. We learn a linear \emph{concept classifier} without bias, parametrized by a weight matrix $\theta \in \mathbb{R}^{\mathcal{|C|} \times d}$ followed by a softmax activation function, returning $p_{\theta}(c'|s_c(x)), \forall c' \in [1,C]$.

The concept classifier is trained by minimizing cross-entropy, where the label of each excerpt $s_c(x)$ the index of the concept for which it was extracted (i.e. $c$):
\begin{flalign}
\mathcal{L}^{concept}(x, \theta, \alpha, \gamma)&=\mathbb{E}_{\bm{s}(x) \sim p_\gamma}[\mathbb{E}_{\bm{z}(x)\sim p_\alpha}[\mathcal{L}^{concept}(\bm{z}(x),\bm{s}(x),\theta)]]\label{eq:ccloss}\\
&=\mathbb{E}_{\bm{s}(x) \sim p_\gamma}[\mathbb{E}_{\bm{z}(x)\sim p_\alpha}[\sum_{c} -z_c(s_c(x))\log p_{\theta}(c|s_c(x))]], \nonumber
\end{flalign}
where $\bm{s}(x)$ refers to the set of excerpts extracted for each concept: $\bm{s}(x)=\{s_c(x), c=1,...,C\}$. The loss considers concepts that are present in $x$ (i.e. $z_c(s_c(x))=1$). The role of the concept classifier is to further enforce concepts consistency and to prevent overlap of concepts, and is jointly train with the rest of the model.

Note that adding a sparsity constraint on the number of concepts present in the inputs enforces semantic consistency, as sparse coding has been shown to induce useful, interpretable representations \citep{Bengio:etal:2013b,Mairal:etal:2010}. We experimentally demonstrate that a sparsity constraint is not sufficient, and can harm output prediction performance. On the opposite, with our concept classifier, sparsity is encouraged as $\mathcal{L}^{concept}$ depends on the concepts that are present, but if concepts are consistent and separable $\mathcal{L}^{concept}$ can be low without harming task performance.

\subsection{EDUCE objective function and optimization}
\label{subsec:obj}
We jointly learn the \emph{concept classifier} and \emph{output classifier} and our objective function is the sum Equations \ref{eq:ocloss} and \ref{eq:ccloss}:
\begin{flalign}
\mathcal{L}(x)&=\mathbb{E}_{\bm{s}(x) \sim p_\gamma}[\mathbb{E}_{\bm{z}(x)\sim p_\alpha}[\mathcal{L}^{output}(y,\bm{z}(x),\delta)+\lambda\mathcal{L}^{concept}(\bm{z}(x),\bm{s}(x),\theta)]] \label{eq:obj}
\end{flalign}
where $\lambda$ controls the strength of the concept consistency constraint wrt the output prediction. The loss $\mathcal{L}(x)$ is differentiable, therefore the gradients with respect to the output classifier's weights ($\delta$) and concept classifier's weights ($\theta$) can be computed and back-propagated. However, the gradients with respect to the parameters of the excerpts extraction ($\gamma$) and the Bernoulli distribution over presence of concepts ($\alpha$) pose an issue due to the sampling of these discrete random variables in Steps 1 and 2. As the explicit computation of the expectation involves expensive summations over all possible values of $\bm{s}(x)$ and $\bm{z}(x)$, we resort to Monte-Carlo approximations of the gradient \citep{Sutton:1998:IRL:551283}, with the loss $\mathcal{L}(x)$ used as reinforcement signal. The gradient derivation is provided in Supplementary Material. Our code will be released upon acceptance.

\section{Related work}
\label{sec:related}

\textbf{Topic models.} Our work relates to topic models, especially to Latent Dirichlet Allocation (LDA) \citep{Blei:etal:2013} where each document is modeled as a mixture of topics and each word in a document relates to one or more topics. While LDA has been recently combined with recurrent neural networks \citep{Zaheer:etal:2015}, the methodology remains different. LDA learns the parameters of a probabilistic graphical model of text generation while we aim at solving a supervised problem in an explanable manner that relies on binary presence/absence of concepts.

\textbf{A posteriori explanations.} Some methods interpret an already trained model, typically using perturbation and gradient-based approach. The most famous method is LIME \citep{Ribeiro:etal:2016}, but other exist \citep{Bach:etal:2015,Shrikumar:etal:2017,Simonyan:etal:2014,Sundararajan:etal:2017}. \citet{Melis:Jaakola:2017} design a model that detects input-output pairs that are causally related. \citet{Kim:etal:2018} learn concept activation vectors. However, the classifier is fixed and concepts are predefined from human annotations, while we learn the concepts unsupervisingly and end-to-end.

\textbf{Self-interpretable models.} Contrarily to the previous line of work, our work falls in the domain of self-interpretable models. Such models produce an explanation simultaneously with their prediction. Related to our work, \citep{Lei:etal:2016,Yu:etal:2019,Bastings:etal:2019} develop interpretable models for NLP tasks by selecting rationales, i.e. parts of text, on which a consequent model bases its prediction. Importantly, a rationale acts as a justification supporting the prediction and is different from our definition of excerpt. In our model, excerpts support the attribution of presence of concepts, and this attribution supports the prediction. \citet{Goyal:etal:2019} propose visual explanations of a classifier's decision and \citet{Alaniz:Akata:2019} use an architecture composed of an observer and a classifier, whose prediction can be exposed as a binary tree. However, contrarily to ours, their model does not provide a local explanation of the decision based on parts of the input. \citet{Melis:Jaakkola:2018} learn a self-explaining classifier that takes as input a set of concepts extracted from the original input. While they define a set of desiderata for what is an interpretable concept, they simply represent extracted concepts as an encoding of the input learned with an auto-encoding loss. \citet{Quint:etal:2018} extend a classic variational auto-encoder architecture with a differentiable decision tree classifier. However, their methodology is different and they only experiment on image data.

\section{Text classification experiments}
\label{sec:expe1}

We experiment on three text classification datasets. The \textbf{DBpedia} ontology classification dataset \citep{Zhang:LeCun:2015b} was constructed by picking $14$ non-overlapping categories from DBpedia 2014 \citep{Lehmann:etal:2014}. There are $14$ classes. We use $56,000$ examples of the train dataset for training, and $56,000$ for validation. For testing, we use $7,000$ examples of the test dataset (using stratified sampling). Also in \citet{Zhang:LeCun:2015b}, the \textbf{AGNews} topic classification dataset was constructed from the AG dataset's 4 largest categories. There are $4$ classes. We separate the training set into $84,000$ training samples and $24,000$ validation samples and report results on the full test dataset. We also experiment on the Stanford Sentiment Treebank (\textbf{SST}) \citep{Socher:etal:2013} that includes 5 sentiment classes. For DBPedia and AGNews we use fixed word vectors trained on Common Crawl \citep{Grave:etal:2018}. For SST, we use fixed Glove word vectors \citep{Pennington:etal:2014}.

\textbf{Comparative models.} Since we are concerned with building self-interpretable models, we do not compare with the methods presented in Section \ref{sec:related} that explain a model a posteriori. Rather, for comparison, we report the performance of (i) using a Bidirectional LSTM on the full text and predict directly from its hidden state (Baseline) (ii) the No Concept Loss model that uses a binary discrete intermediate representation but no concept classifier, it corresponds to $\lambda=0$ and its training objective is simply $\mathcal{L}^{output}$ and (iii) adding a $L_1$-norm sparsity constraint to the No Concept Loss model (referred to as No Concept Loss + $L_1$), i.e. Equation \ref{eq:obj} changes to:
\begin{flalign}
\mathcal{L}(x)&=\mathbb{E}_{s(x) \sim p_\gamma}[\mathbb{E}_{\bm{z}(x)\sim p_\alpha}[\mathcal{L}^{output}(y,\bm{z}(x),\delta) + \lambda_{L_1}|\bm{z}|]].
\end{flalign}
Therefore both No Concept Loss and No Concept Loss + $L_1$ models extract excerpts and use a binary intermediate representation, but are trained without concept loss.

\textbf{Hyperparameters selection.} We tried different values of hyperparameters (details in supplementary Section \ref{sec:detailssupp}) and for each set of hyperparameters, we run $5$ different random seeds. We choose the best hyperparameters set using its average validation performance over the $5$ random seeds.

\textbf{Metrics.} We report the test output accuracy (Output Acc.) over the task. Evaluating interpretability of concepts is challenging. For EDUCE, we can compute the accuracy of the concept classifier on the test data. However, this should be low for models No Concept Loss and No Concept Loss + $L_1$ that do not train with a concept loss. Therefore, to evaluate interpretability, we report an \emph{a posteriori} concept accuracy: after training, for each model (including EDUCE), we gather the excerpts extracted ($z_c=1$) in the test data. We separate these excerpts into two sets (training and testing, note that these are both generated from the test data). We train a new, separate concept classifier \emph{a posteriori} on the extracted excerpts to evaluate and we report its performance as a posteriori concept accuracy (A Posteriori Concept Acc.).
\begin{figure}[t!]
  \begin{minipage}[t]{0.50\textwidth}
\vspace{0pt} \centering
\setlength\tabcolsep{0.6pt}
\small{
\begin{tabular}{+c|+c|^c|^c}
\hline
Data & Model & Output Acc. (\%) & \makecell{A Posteriori \\ Concept Acc. (\%)} \\ \cline{2-4}
\hline
{\multirow{4}{*}{\rotatebox[origin=c]{90}{DBPedia}}} & EDUCE & $97.0\pm0.1$& $\mathbf{82.4\pm0.8}$\\
\cline{2-4}
~&No Concept Loss&$97.4\pm0.1$&  $25.9\pm0.6$\\
\cline{2-4}
~&No Concept Loss+$L_1$& $96.5\pm0.2$& $44\pm2.6$\\ \cline{2-4}
~&Baseline& $\mathbf{98.75\pm0.0}$  &  n/a \\ \hline
\hline
{\multirow{4}{*}{\rotatebox[origin=c]{90}{AGNews}}} & EDUCE & $87.5\pm0.2$& $\mathbf{78\pm6.5}$\\
\cline{2-4}
~&No Concept Loss&$88.2\pm0.1$ & $31.0\pm0.7$\\
\cline{2-4}
~&No Concept Loss+$L_1$& $86.3\pm0.7$& $56\pm3.2$\\
\cline{2-4}
~&Baseline & $\mathbf{92.08\pm0.1}$  & n/a \\
\hline
\end{tabular}
}
\captionof{table}{Test performance on DBPedia and AGNews (mean $\pm$ SEM).}
\label{tab:res}
\end{minipage}
\hfill
\begin{minipage}[t]{0.42\textwidth}
\vspace{0pt} \centering
  \includegraphics[width=\textwidth]{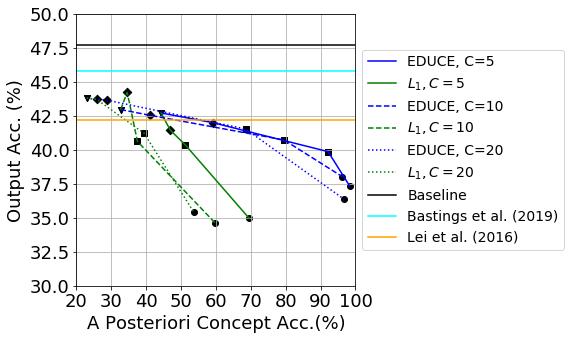}
  \captionof{figure}{SST test performance, output accuracy vs a posteriori concept accuracy. A posteriori concept accuracy is not applicable for Baseline, \citet{Lei:etal:2016} and \citet{Bastings:etal:2019}.}
  \label{fig:resSST}
\end{minipage}
\vspace{-5mm}
\end{figure}

\textbf{Quantitative results on DBPedia and AGNews.} Table \ref{tab:res} reports test performance on the DBPedia and AGNews datasets for $C=10$ concepts. We report results with values $\lambda=0.1$ and $\lambda_{L_1}=0.1$ that give best trade-off performance between output and concept accuracies. Complete results are available in supplementary Table \ref{tab:allres}. For all metrics we report the mean and standard error of the mean (SEM) over the training random seeds. Table \ref{tab:res} shows that the Baseline model outperforms the other models that all employ a binary discrete encoding of the input. This is expected, and shows the trade-off between interpretability and output accuracy. In terms of output accuracy, EDUCE is comparable with its counterpart that does not encourage concept consistency (No Concept Loss), loosing only $0.4\%$ on DBPedia and $0.7\%$ on AGNews, and as expected outperforms it in terms of concept accuracy. Adding an $L_1$ constraint to the No Concept Loss model increases the consistency of the concepts (as per A posteriori Concept Accuracy values) yet is largely outperformed by EDUCE on that metric, and output accuracy is also lower than EDUCE's.

\textbf{Interpreting EDUCE on AGNews.} We turn to show how EDUCE's output prediction is easily interpretable. The following results were generated with $\lambda=0.1$ and $C=10$ concepts. Figure \ref{tab:ex1} shows examples of the AGNews test set that were correctly classified by EDUCE. The underlined words correspond to the excerpts extracted for different concepts. Separately, in Figure \ref{tab:ex2} we show, for each concept detected in the examples of Figure \ref{tab:ex1}, examplar excerpts extracted from others test documents (excerpts are separated by ``/"). We interpret the concepts' meaning as follows: concept 0 maps to informatics notions, concept 2 to corporations/petrol, concept 4 to the notions of investments and finance, concept 6 to sports events and concept 8 to governmental/state affaires. Concept 1 is less clearly defined. Note that in Figure \ref{tab:ex2} excerpts are consistent yet are extracted in texts from  multiple output classes. The excerpts extracted in the examples of Figure \ref{tab:ex1} are consistent with these interpretations. These results show how easily the classification of any text can be explained by the detection of multiple, relevant, and intelligible concepts. More qualitative examples are in supplementary Section \ref{sec:expesupp}.

\begin{figure}[h!]
\begin{subtable}[t]{\textwidth}
\small{
\begin{tabularx}{\textwidth}{X}
  Class Business:
  \emph{ \colorunderline[color=RawSienna,above=0.5pt]{\text{sports}} \colorunderline[color=RawSienna,above=0.5pt]{ \colorunderline[color=LimeGreen,above=0.5pt]{\text{retailer}}} \colorunderline[color=RawSienna,above=0.5pt]{ \colorunderline[color=LimeGreen,above=0.5pt]{\text{jjb}}} \colorunderline[color=LimeGreen,above=0.5pt]{\text{yesterday}} reported a near 25 drop in profits and continuing poor sales , and \colorunderline[color=Magenta,above=0.5pt]{\text{ended}} \colorunderline[color=Magenta,above=0.5pt]{\text{shareholders}} \colorunderline[color=Magenta,above=0.5pt]{\text{\#39}} \colorunderline[color=Magenta,above=0.5pt]{\text{hopes}} \colorunderline[color=Magenta,above=0.5pt]{\text{of}} \colorunderline[color=Magenta,above=0.5pt]{\text{a}} \colorunderline[color=Magenta,above=0.5pt]{\text{takeover}} by announcing that a potential bidder had walked away .}\\ \hline
    Class World:
  \emph{ \colorunderline[color=Purple,above=0.5pt]{\text{bangkok}} \colorunderline[color=Purple,above=0.5pt]{\text{,}} \colorunderline[color=Purple,above=0.5pt]{\text{thailand}} sept . 30 , 2004 - millions of volunteers led by \colorunderline[color=RawSienna,above=0.5pt]{\text{emergency}} \colorunderline[color=RawSienna,above=0.5pt]{\text{teams}} \colorunderline[color=RawSienna,above=0.5pt]{\text{fanned}} out across thailand on thursday in a new drive to fight bird flu after the prime \colorunderline[color=Orange,above=0.5pt]{\text{minister}} \colorunderline[color=Orange,above=0.5pt]{\text{gave}} \colorunderline[color=Orange,above=0.5pt]{\text{officials}} 30 days to eradicate the epidemic .}\\ \hline
  Class Sports:
  \emph{ the \colorunderline[color=Orange,above=0.5pt]{\text{spanish}} \colorunderline[color=Orange,above=0.5pt]{\text{government}} \colorunderline[color=Orange,above=0.5pt]{\text{responded}} to diplomatic pressure from britain yesterday by starting a search for fans who racially abused england players during a quot friendly quot \colorunderline[color=RawSienna,above=0.5pt]{\text{football}} \colorunderline[color=RawSienna,above=0.5pt]{\text{match}} \colorunderline[color=RawSienna,above=0.5pt]{\text{with}} spain .}\\ \hline
  Class Sci/Tech:
\emph{ los angeles ( reuters ) - a group of technology companies including texas instruments inc . \&lt txn . n\&gt , stmicroelectronics \&lt stm . pa\&gt and broadcom \colorunderline[color=LimeGreen,above=0.5pt]{\text{corp}} \colorunderline[color=LimeGreen,above=0.5pt]{\text{.}} \colorunderline[color=LimeGreen,above=0.5pt]{\text{\&lt}} \colorunderline[color=LimeGreen,above=0.5pt]{\text{brcm}} . o\&gt , on thursday said they will propose a \colorunderline[color=Gray,above=0.5pt]{\text{new}} \colorunderline[color=Gray,above=0.5pt]{ \colorunderline[color=Blue,above=0.5pt]{\text{wireless}}} \colorunderline[color=Gray,above=0.5pt]{ \colorunderline[color=Blue,above=0.5pt]{\text{networking}}} \colorunderline[color=Blue,above=0.5pt]{\text{standard}} up to 10 times the speed of the current generation .}\\ \hline
\end{tabularx}
}
\caption{AGNews test examples correctly classified by EDUCE. Underlined set of words are excerpts extracted, one color per concept.}
\label{tab:ex1}
\end{subtable}
\begin{subtable}[t]{\textwidth}
  \vspace{2mm}
\small{
\begin{tabularx}{\textwidth}{|>{\centering\arraybackslash}p{1.3cm}|X|}
  \hline
  \textcolor{Blue}{Concept 0} &
  software services giant
  /
  moonwalk to home
  /
  video display chip
  /
  downloading music .
  \\ \hline
   \textcolor{Gray}{Concept 1} &
  upcoming my prerogative video
  /
  his ever-growing swimming
  /
  launch of a video display chip
  /
  illegality of downloading
  \\ \hline
   \textcolor{LimeGreen}{Concept 2} &
  oil market .
  /
  oil giant sibneft
  /
  oil prices and
  /
  corp . \&lt brcm
 \\ \hline
 \textcolor{Magenta}{Concept 4} &
 cash settlement of up to \#36 50 million
 /
 six-year deal worth about \$40 million
 /
 the dollar dipped to a four-week low against the euro
 /
 five shares ,
 \\ \hline
 \textcolor{RawSienna}{Concept 6} &
olympic 100-meter freestyle
/
sox ' family
/
olympics should help
/
athletes were already
\\ \hline
 \textcolor{Orange}{Concept 8} &
frail pope john paul
/
indian army major shot
/
unions representing workers
/
goverment representatives .
\\ \hline
\end{tabularx}
}
\captionof{figure}{Examples of excerpts that are extracted \textbf{accross the test set}, corresponding to the concepts detected in Figure \ref{tab:ex1}. Colors match the colors used in Figure \ref{tab:ex1}.}
\label{tab:ex2}
\end{subtable}
\caption{Interpretation of EDUCE prediction through concept analysis.}
\vspace{-5mm}
\end{figure}

\textbf{SST dataset} Figure \ref{fig:resSST} shows output accuracy versus a posteriori concept accuracy on the SST dataset, for different different number of concepts $C$ and different values of $\lambda$: Each marker is a different value of $\lambda$, and values are (from left to right markers) $\{0,0.01,0.1,0.5\}$. We do the same for the No Concept Loss + $L_1$ model. Note that the left most markers in all curves correspond to $\lambda=0$ and $\lambda_{L_1}=0$, therefore to the model referred to as No Concept Loss. While our goal and model is very different than \citet{Lei:etal:2016, Bastings:etal:2019}, we report their test scores (obtained using $40\%$ of the text on the SST dataset, we report directly from \citet{Bastings:etal:2019}). The a posteriori concept accuracy for Baseline, \citet{Lei:etal:2016} and \citet{Bastings:etal:2019} is not applicable. Figure \ref{fig:resSST} confirms our results: EDUCE (in \textcolor{blue}{blue}) overperforms the model No Concept Loss + $L_1$ (in \textcolor{green}{green}) on both output and a posteriori concept accuracies, and achieve output accuracy comparable to \citet{Lei:etal:2016} (orange line) while having $60\%$ a posteriori concept accuracy.

\begin{figure}[h!]
    \begin{subfigure}[t]{0.66\textwidth}
        \begin{subfigure}[t]{0.49\textwidth}
          \centering
          \includegraphics[width=\textwidth]{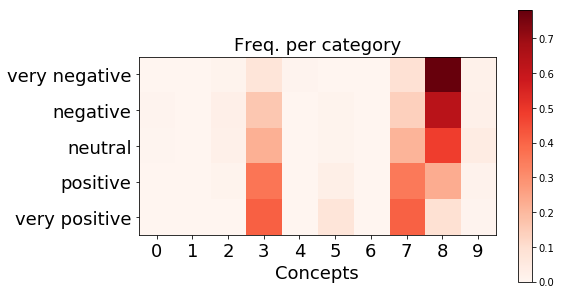}
          \caption{Per class concept frequency.}
          \label{fig:freqsst}
        \end{subfigure}
        \begin{subfigure}[t]{0.49\textwidth}
          \centering
          \includegraphics[width=\textwidth]{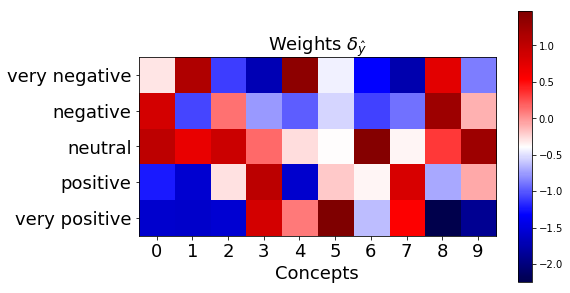}
          \caption{Output classifier weight matrix $\delta$.}
          \label{fig:deltasst}
        \end{subfigure}
  \end{subfigure}
  \begin{subfigure}[t]{0.30\textwidth}
    \begin{subfigure}[t]{\textwidth}
    \centering
      \includegraphics[width=\textwidth]{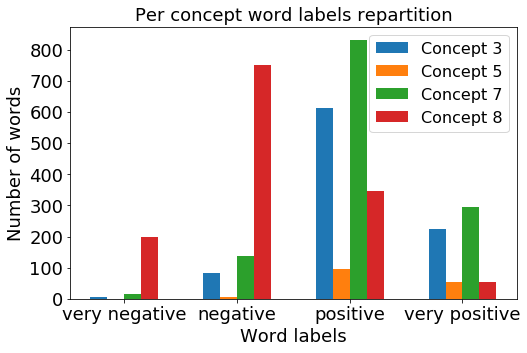}
      \caption{Repartition of words labels for concepts 3, 5, 7 and 8.}
      \label{fig:reptkl}
    \end{subfigure}
  \end{subfigure}
  \caption{Analysis of concepts on SST.}
\end{figure}

Figure \ref{fig:freqsst} shows the empirical frequency of presence of each concept, per output class in the SST dataset, using $\lambda=0.1$ and $C=10$ concepts. We see that concepts 3, 5 and 7 are often triggered when a sample is positive or very positive, and on the opposite concept 8 is present in negative/very negative texts. The values of weight matrix $\delta$ of the output classifier in Figure \ref{fig:deltasst} proves that the presence of these concepts is responsible for the output classifier's prediction. We can further analyze the concepts: Each word in SST sentences is annotated with a label for the sentiment it expresses, referred to as \emph{word label}. We do not use them during training or validation, but at test-time this allows us to analyze the repartition of words labels in the set of excerpts extracted for each concept. Figure \ref{fig:reptkl} shows that indeed the excerpts selected for concepts 3, 5 and 7 are mostly composed of positive or very positive words, and the excerpts extracted when concept 8 is triggered are mostly composed of negative or very negative words\footnote{We do not show the amount of neutral words selected as it squeezes the histograms.}.

\section{Multi aspect Sentiment Analysis experiment}
\label{sec:expe2}

\textbf{BeerAdvocate dataset} While our paradigm is different than \citet{Lei:etal:2016, Bastings:etal:2019}, we also perform a multi-aspect sentiment regression experiment on the pre-processed subset of the BeerAdvocate\footnote{https://www.beeradvocate.com/} dataset \citep{McAuley:etal:2012}. It consists of $260,000$ beer reviews where ratings for four aspects (look, smell, palate and taste) are given, as well as an overall rating. The reviews are separated into training/validation sets specific to each aspect, and ratings are mapped to scalars in $[0, 1]$. This is a regression task, so we modify the Baseline model and EDUCE in the last layer of the prediction with a sigmoid activation function, and use Mean Squared Error (MSE) instead of cross-entropy for $\mathcal{L}^{output}(y,\bm{z}(x),\delta)$. \citet{McAuley:etal:2012} provided a test set ($994$ reviews) with sentence-level rationale annotations: different parts of each review are annotated with one (or multiple) aspect label, indicating what aspect it covers (referred to as gold rationales). Importantly, we do not use these gold rationales during training or validation, but use to evaluate our model. Hyperparameters selection is similar to our classification experiments.

We train EDUCE on the prediction of the 5-values vector at once (4 aspects and the overall score, there are therefore $5$ scalar ratings to predict) using the $260,000$ reviews. We compare EDUCE's performance with the Baseline that accesses the full text. The test MSE is $0.0089\pm0.0$ (mean and SEM accross training seeds) for the  Baseline model and $0.0119\pm0.0$ for EDUCE using $\lambda=0.01$ and $C=10$ concepts, with a corresponding concept accuracy of $92.8\pm0.7$. Of specific interest to us is the per-concept precision of gold rationales and the total percentage of selected words. Indeed, Table \ref{tab:beer1} shows that when predicting the 4 aspects and the overall score, some concepts are capturing a specific aspect: for example, the precision of Concept 7 on the Appearance aspect is $97.09\%$. This is confirmed by Table \ref{tab:beer2} that shows that excerpts extracted for Concept 7 are mostly related to the color and head (the frothy foam on top of beer) of the beer\footnote{Tables \ref{tab:beer1} and \ref{tab:beer2} reports results of the best seed for the cross-validated hyperparameters.}. Supplementary Table \ref{tab:beer4} reports examplar excerpts for all concepts.

\begin{table}[h!]
  \begin{minipage}[t]{0.60\textwidth}
\vspace{0pt} \centering
  \setlength\tabcolsep{1pt}
\small{
\begin{tabularx}{\textwidth}{|>{\centering\arraybackslash}p{1.3cm}|X|}
\hline
Concept 0 &
fruity esters , and/ fruits , caramelized pecans , and/ toffee and caramel accents ,/ coffee and chocolate flavors/ earthy hop resin . \\ \hline
Concept 2 &
creamy and a/ chewy and rich and drinkability/ creamy mouthfeel that/ smooth and just velvety on the/ thick , and \\ \hline
Concept 3 &
rich malt scents/ aroma is quite hoppy with big citrus/ tons of different sweet malts , toffe/ smells extremely roasty/ boom of grapefruity \\ \hline
Concept 7 &
good head and lacing/ beautiful golden-amber color/ creamy tan head/ nice deep brown color/ proud head has settled . nothing \\ \hline
Concept 9 &
carbonation is graceful/ drinkability is excellent/ mouthfeel is wonderful/ mouthfeel is exemplary/ drinkability : excellent \\
\hline
\end{tabularx}
}
\caption{Examples of excerpts extracted on the Beer test set.}
\label{tab:beer2}
\end{minipage}
\begin{minipage}[t]{0.40\textwidth}
\vspace{0pt}  \centering
    \setlength\tabcolsep{1pt}
\small{
\begin{tabular}{|c|^c|^c|^c|^c|}
\hline
Concept & Appearance & Smell & Palate & Taste \\
\hline
Concept 0&0.85 &\textbf{55.75} &1.79 &40.60  \\ \hline
Concept 1&0.54 &5.48 &11.70 &30.58  \\ \hline
Concept 2&1.22 &0.90 &\textbf{78.18} &15.03  \\ \hline
Concept 3&1.46 &\textbf{92.39} &0.30 &5.34  \\ \hline
Concept 4&45.33 &0.23 &0.26 &0.16  \\ \hline
Concept 5&7.27 &22.50 &2.27 &28.86  \\ \hline
Concept 6&0.36 &5.86 &6.40 &28.73  \\ \hline
Concept 7&\textbf{97.09} &0.98 &0.45 &0.53  \\ \hline
Concept 8&0.18 &0.09 &18.07 &5.93  \\ \hline
Concept 9&0.00 &0.00 &\textbf{68.42} &10.53  \\ \hline
\end{tabular}
}
\caption{Per-concept precision of gold rationales (in \% for each aspect), trained to predict all 4 aspects and the overall score. In bold we emphasize the concepts which precision is above $50\%$ for an aspect's gold rationales.}
\label{tab:beer1}
\end{minipage}
\end{table}
As in \citet{Lei:etal:2016, Bastings:etal:2019} we also train EDUCE on the prediction of each of the first three aspects separately. Supplementary Table \ref{tab:beer3} reports the precision of gold rationales for each aspect. While the results are less convincing than the models specifically designed for this task, extracted excerpts match gold rationales to a certain extent.

\section{Discussion and perspectives}
We propose a new self-interpretable model, EDUCE, that maps inputs to an easy-to-interpret representation on which the output prediction relies. We experimentally demonstrate its ability (i) to extract semantically meaningful concepts described by consistent excerpts, and (ii) to ground its output prediction on the presence/absence of these concepts in each input. EDUCE generates relevant explanations while retaining high predictive performance. As future direction for research, we contemplate using hierarchical representations, as human notions can often be represented with hierarchical structures, for example as in WordNet \citep{Miller:1995}. Furthemore, our principles can be applied on different types of inputs like images, and we provide initial experiment in Supplementary Section \ref{imludo}.

\bibliography{interpretable}
\bibliographystyle{iclr2020_conference}
\appendix
\section{Details on the learning algorithm}

As mentioned in our main paper, when computing Equation \ref{eq:obj} the cross-entropy losses of the output classifier and concept classifier are differentiable, therefore the gradients with respect to the weights $\delta$ and $\theta$ can be computed. However, the gradients with respect to the parameters of the LSTM ($\gamma$) and the Bernouilli distribution over presence of concepts ($\alpha$) pose a difficulty because we sample $\forall c~s_c(x) \sim p_{\gamma}(s|x,c)$ and $\forall c~z_c \sim p_{\alpha}(z_c|s_c(x),c)$.

We use straight-through estimator \citep{Bengio:etal:2013} in the backward pass for the parameters $\alpha$, and resort to Monte-Carlo approximations of the gradient employed in Reinforcement Learning \citep{Sutton:1998:IRL:551283} for the parameters $\gamma$. The loss $\mathcal{L}(x)$ serves as reinforcement signal. We weight the two terms of the gradient using a parameter $r$ as follows:
\begin{flalign}
  \nabla_{\delta,\theta,\gamma,\alpha} \mathcal{L}(x) \widehat{=} (1-r)\nabla_{\delta,\theta,\alpha} \mathcal{L}(x) + r \mathbb{E}_{s(x) \sim p_\gamma}[(\mathcal{L}(x) - b)\nabla_{\gamma}\log p_{\gamma} (s(x)|x)]
\end{flalign}
where $r$ controls the strength of the Reinforcement Learning term and $b$ is the average of the loss, used as control variate.

\section{Experimental details and detailed results.}

We monitor validation performance and use early stopping. For the comparative models we monitor the output prediction accuracy. For EDUCE we monitor the sum of the output prediction accuracy and the concept classifier accuracy (weighted by the parameter $\lambda$).

\subsection{Hyperparameters considered}
\label{sec:detailssupp}
In all experiment, we increase the value of $\lambda$ by $\10\%$ every epoch. In order to choose the best set of hyperparameters, we take the average value on the $5$ random seeds.

\textbf{DBPedia and AGNews}
We try the following ranges of hyperparameters $\lambda \in \{0.0,0.001,0.01, 0.1\}$, $\lambda_{L1} = \{0.0,0.001, 0.01, 0.1, 1\}$, $r \in \{0.01, 0.1\}$, learning rate $lr=\{0.0005,0.001\}$. The hidden state of the LSTM is of size $200$. We use Adam optimizer \citep{Kingma:and:Ba:2014}, batches of size $64$. For the Baseline model, we try adding weight decay in $\{0.0,1e-6\}$ and dropout in $\{0.0,0.1,0.2\}$.

\textbf{SST}
We try the following ranges of hyperparameters $\lambda \in \{0.0,0.001,0.01, 0.1\}$, $\lambda_{L1} = \{0.0,0.001, 0.01, 0.1, 1\}$, $r \in \{0.01, 0.1\}$, learning rate $lr=\{0.0002,0.001\}$ and weight decay in $\{0.0,1e-6\}$. The hidden state of the LSTM is of size $150$. We use Adam optimizer \citep{Kingma:and:Ba:2014}, batches of size $25$. For the Baseline model, we follow \citet{Bastings:etal:2019} and use weight decay $1e-6$ and dropout $\{0.5\}$.

\textbf{BEER}
We try the following ranges of hyperparameters $\lambda \in \{0.001,0.01, 0.1\}$, $r \in \{0.001,0.01, 0.1\}$, learning rate $lr=\{0.0005,0.001\}$. The hidden state of the LSTM is of size $200$. We use Adam optimizer \citep{Kingma:and:Ba:2014}, batches of size $256$. We also clip the gradient to the value of $1$ and employ an entropy maximization regularization on mean of the Bernoulli distributions $\forall c, p_{\alpha}(z_c|s_c(x),c)$, weighted by $0.001$. We train all models for $250$ epochs and use early stopping on the validation set. For the Baseline model on the BEER dataset, we use the same hyperparameters as in \citet{Bastings:etal:2019} ($lr=0.0004$, hidden dimension $200$, dropout in $\{0.1,0.2\}$ and weight decay $1e-6$) and train for $100$ epochs.

\subsubsection{Detailed results on AGNews and DBPedia}
\label{sec:detailedres}

Table \ref{tab:allres} details output accuracy, a posteriori concept accuracy, and sparsity (number of concepts presents per input) for different values of $\lambda$ and $\lambda_{L_1}$ considered.
\begin{table}[h!]
\centering
\setlength\tabcolsep{1pt} 
\small{
\begin{tabular}{+c|+c|^c|^c|^c}
Dataset& Model & Output Acc. (\%) & \makecell{A Posteriori \\ Concept Acc. (\%)} & Sparsity \\
{\multirow{8}{*}{\rotatebox[origin=c]{90}{DBPEDIA}}} & Baseline& $98.75\pm0.0$  &  n/a &  n/a \\
\cline{2-5}
~&No Concept Loss & $97.4\pm0.1$& $25.9\pm0.6$& $5.51\pm0.1$\\
\cline{2-5}
~& EDUCE $\lambda=0.001$ & $97.3\pm0.1$& $28\pm1.7$& $5.60\pm0.1$\\
\cline{2-5}
~& EDUCE $\lambda=0.01$ & $97.3\pm0.1$& $39\pm3.5$& $5.39\pm0.1$\\
\cline{2-5}
~& EDUCE $\lambda=0.1$ & $97.0\pm0.1$& $82.4\pm0.8$& $3.5\pm0.2$\\
\cline{2-5}
\cline{2-5}
~&No Concept Loss + $L_1$, $\lambda_{L_1}=0.001$ & $97.4\pm0.1$& $27\pm1.9$& $5.5\pm0.1$\\
\cline{2-5}
~&No Concept Loss + $L_1$, $\lambda_{L_1}=0.01$& $97.46\pm0.1$& $28\pm2.1$& $5.12\pm0.0$\\
\cline{2-5}
~&No Concept Loss + $L_1$, $\lambda_{L_1}=0.1$& $96.5\pm0.2$& $44\pm2.6$& $3.1\pm0.1$\\
\cline{2-5}
~&No Concept Loss + $L_1$, $\lambda_{L_1}=1.00$&$84\pm2.0$& $77\pm2.4$& $1.31\pm0.0$\\
\hline
\hline
{\multirow{8}{*}{\rotatebox[origin=c]{90}{AGNews}}} & Baseline& $92.08\pm0.1$  & n/a & n/a \\
\cline{2-5}
~&No Concept Loss & $88.2\pm0.1$& $31.0\pm0.7$& $5.2\pm0.2$\\
\cline{2-5}
~& EDUCE $\lambda=0.001$ & $88.4\pm0.1$& $33\pm1.6$& $5.2\pm0.2$\\
\cline{2-5}
~& EDUCE $\lambda=0.01$ & $88.6\pm0.1$& $40\pm2.1$& $4.6\pm0.2$\\
\cline{2-5}
~& EDUCE $\lambda=0.1$ & $87.5\pm0.2$& $78\pm6.5$& $2.4\pm0.2$\\
\cline{2-5}
~&No Concept Loss + $L_1$, $\lambda_{L_1}=0.001$ & $88.3\pm0.2$& $31.6\pm0.7$& $5.1\pm0.2$\\
\cline{2-5}
~&No Concept Loss + $L_1$, $\lambda_{L_1}=0.01$& $88.7\pm0.2$& $34\pm1.4$& $4.6\pm0.2$\\
\cline{2-5}
~&No Concept Loss + $L_1$, $\lambda_{L_1}=0.1$& $86.3\pm0.7$& $56\pm3.2$& $2.2\pm0.1$\\
\cline{2-5}
~&No Concept Loss + $L_1$, $\lambda_{L_1}=1.00$& $53\pm3.2$& $89\pm2.8$& $0.50\pm0.0$\\
\hline
\end{tabular}
}
\caption{Test performance on DBPedia and AGNews (mean $\pm$ SEM) for all values of $\lambda$ and $\lambda_{L_1}$.}
\label{tab:allres}
\end{table}

\subsection{Additional results on Beer}

\begin{table}[h!]
\centering
  \setlength\tabcolsep{1pt}
\small{
\begin{tabularx}{\textwidth}{|>{\centering\arraybackslash}p{1.3cm}|X|}
\hline
Concept 0 &
fruity esters , and/ fruits , caramelized pecans , and/ toffee and caramel accents ,/ coffee and chocolate flavors/ earthy hop resin . \\ \hline
Concept 1 &
delicious and with/ enjoyed this beer/ damn good beer/ nice stout that/ good beer to \\ \hline
Concept 2 &
creamy and a/ chewy and rich and drinkability/ creamy mouthfeel that/ smooth and just velvety on the/ thick , and \\ \hline
Concept 3 &
rich malt scents/ aroma is quite hoppy with big citrus/ tons of different sweet malts , toffe/ smells extremely roasty/ boom of grapefruity \\ \hline
Concept 4 &
octoberfest : say the word and you probably/ 2004 vintage : pours/ 12 ounce bottle acquired in a trade/ i hate to be the guy/ 12oz bottle , best before 9/09 , sampled 5/16/09 \\ \hline
Concept 5 &
which is fantastic/ it is delicious/ is intensely packed/ and a lovely/ builds a wonderful \\ \hline
Concept 6 &
'm really digging/ a damn good/ a super fresh growler/ the better abitas/ a great traditional \\ \hline
Concept 7 &
good head and lacing/ beautiful golden-amber color/ creamy tan head/ nice deep brown color/ proud head has settled . nothing \\ \hline
Concept 8 &
easy to drink/ hot summer day/ drinkability easy/ easy to down/ can easily turn this 'strong ale ' into a session
\\ \hline
Concept 9 &
carbonation is graceful/ drinkability is excellent/ mouthfeel is wonderful/ mouthfeel is exemplary/ drinkability : excellent \\
\hline
\end{tabularx}
}
\caption{Examples of excerpts extracted on the Beer test set.}
\label{tab:beer4}
\end{table}

As in \citet{Lei:etal:2016, Bastings:etal:2019}, we also train EDUCE on the prediction of each of the first three aspects (the 4th aspect they do not experiment on)\footnote{\citet{Bastings:etal:2019} call the third aspect Taste but they actually refer to Palate.}. Table \ref{tab:beer3} reports the precision of gold rationales for that aspect and the total percentage of selected words, values for \citet{Lei:etal:2016, Bastings:etal:2019} are taken from \citet{Bastings:etal:2019}. We use $C=10$ and $\lambda=0.1$. We consider a word as selected if any of the concept selects it, and count it as one selected even if it is selected by multiple concepts. While the results are less convincing than the aforementioned rationales models that are specifically designed for this task, we see that extracted excerpts match gold rationales to a certain extent, especially for the aspects Appearance and Smell. For the Palate aspect, manual inspection shows us that for some seeds, the model defines a concept dedicated to the notion of numbers (e.g. corresponding to excerpts ``12 ounce bottle'') that is consistent, yet not part of the gold rationale.

\begin{table}[t!]
\centering
\setlength\tabcolsep{1pt}
\small{
\begin{tabular}{|c|c|c|c|c|c|c|}
\hline
Model & \multicolumn{2}{c|}{Appearance} & \multicolumn{2}{c|}{Smell} & \multicolumn{2}{c|}{Palate} \\\hline
~ & Prec. & Extr. & Prec. & Extr.& Prec. & Extr. \\
\hline
EDUCE & $81\pm6.9$& $4.9\pm0.5$ & $79\pm3.1$& $4.8\pm0.4$ & $50\pm6.6$& $4.1\pm0.6$\\
\citet{Lei:etal:2016}& 96.3 & 14 & 95.1 & 7 & 80.2 & 7\\
\citet{Bastings:etal:2019}& 98.1 & 13 & 96.8 & 7 & 89.8 &7\\
\hline
\end{tabular}
}
\caption{Precision (Prec., in \%) of gold rationales and percentage of extracted text (Extr. in \%) when trained separately on each aspect.}
\label{tab:beer3}
\end{table}

\section{Additional qualitative examples on DBPedia and AGNews}
\label{sec:expesupp}

\begin{table}[t!]
\small{
\begin{tabularx}{\textwidth}{X}
Class Sports:
\emph{ just imagine what david ortiz could do on a good night ' s rest . ortiz spent the night before last with his baby boy , d ' angelo , who is barely 1 month old . he had planned on attending \colorunderline[color=Salmon,above=0.5pt]{\text{the}} \colorunderline[color=Salmon,above=0.5pt]{\text{red}} \colorunderline[color=RawSienna,above=0.5pt]{ \colorunderline[color=Salmon,above=0.5pt]{\text{sox}}} \colorunderline[color=RawSienna,above=0.5pt]{ \colorunderline[color=Salmon,above=0.5pt]{\text{'}}} \colorunderline[color=RawSienna,above=0.5pt]{ \colorunderline[color=Salmon,above=0.5pt]{\text{family}}} \colorunderline[color=Salmon,above=0.5pt]{\text{day}} \colorunderline[color=Salmon,above=0.5pt]{\text{at}} fenway park yesterday morning , but he had to sleep in . after all , ortiz had a son at home , and he . . .}\\ \hline
Class Sports:
\emph{ foxborough -- looking at his ridiculously developed upper body , with huge biceps and hardly an ounce of fat , it ' s easy to see why ty law , arguably the best \colorunderline[color=RawSienna,above=0.5pt]{\text{cornerback}} \colorunderline[color=RawSienna,above=0.5pt]{\text{in}} \colorunderline[color=RawSienna,above=0.5pt]{\text{football}} , chooses physical play over finesse . that ' s not to imply that he ' s lacking a finesse component , because he can shut down his side of \colorunderline[color=Salmon,above=0.5pt]{\text{the}} \colorunderline[color=Salmon,above=0.5pt]{\text{field}} \colorunderline[color=Salmon,above=0.5pt]{\text{much}} \colorunderline[color=Salmon,above=0.5pt]{\text{as}} \colorunderline[color=Salmon,above=0.5pt]{\text{deion}} \colorunderline[color=Salmon,above=0.5pt]{\text{sanders}} \colorunderline[color=Salmon,above=0.5pt]{\text{.}} \colorunderline[color=Salmon,above=0.5pt]{\text{.}} \colorunderline[color=Salmon,above=0.5pt]{\text{.}}}\\ \hline
Class Sports:
\emph{ \colorunderline[color=Purple,above=0.5pt]{\text{ap}} \colorunderline[color=Purple,above=0.5pt]{\text{-}} \colorunderline[color=Purple,above=0.5pt]{\text{american}} \colorunderline[color=Purple,above=0.5pt]{\text{natalie}} coughlin won olympic gold in \colorunderline[color=Salmon,above=0.5pt]{\text{the}} \colorunderline[color=RawSienna,above=0.5pt]{ \colorunderline[color=Salmon,above=0.5pt]{\text{100-meter}}} \colorunderline[color=RawSienna,above=0.5pt]{ \colorunderline[color=Salmon,above=0.5pt]{\text{backstroke}}} \colorunderline[color=RawSienna,above=0.5pt]{ \colorunderline[color=Salmon,above=0.5pt]{\text{monday}}} \colorunderline[color=Salmon,above=0.5pt]{\text{night}} \colorunderline[color=Salmon,above=0.5pt]{\text{.}} coughlin , the only woman ever to swim under 1 minute in the event , finished first in 1 minute , 0 . 37 seconds . kirsty coventry of zimbabwe , who swims at auburn university in alabama , earned the silver in 1 00 . 50 . laure manaudou of france took bronze in 1 00 . 88 .}\\ \hline
Class World:
\emph{ \colorunderline[color=Purple,above=0.5pt]{\text{lourdes}} \colorunderline[color=Purple,above=0.5pt]{\text{,}} \colorunderline[color=Purple,above=0.5pt]{\text{france}} - a \colorunderline[color=Orange,above=0.5pt]{\text{frail}} \colorunderline[color=Orange,above=0.5pt]{\text{pope}} \colorunderline[color=Orange,above=0.5pt]{\text{john}} \colorunderline[color=Orange,above=0.5pt]{\text{paul}} ii , breathing heavily and gasping at times , celebrated an open-air mass on sunday for several hundred thousand pilgrims , many in wheelchairs , at a shrine to the virgin mary that is associated with miraculous cures . at one point he said help me in polish while struggling through his homily in french . . .}\\ \hline
Class World:
\emph{ najaf \colorunderline[color=Purple,above=0.5pt]{\text{,}} \colorunderline[color=Purple,above=0.5pt]{\text{iraq}} \colorunderline[color=Purple,above=0.5pt]{\text{-}} explosions and \colorunderline[color=Red,above=0.5pt]{\text{gunfire}} \colorunderline[color=Red,above=0.5pt]{\text{rattled}} \colorunderline[color=Red,above=0.5pt]{\text{through}} \colorunderline[color=Red,above=0.5pt]{\text{the}} city of najaf as u . s . \colorunderline[color=Orange,above=0.5pt]{\text{troops}} \colorunderline[color=Orange,above=0.5pt]{\text{in}} \colorunderline[color=Orange,above=0.5pt]{\text{armored}} \colorunderline[color=Orange,above=0.5pt]{\text{vehicles}} and tanks rolled back into the streets here sunday , a day after the collapse of talks - and with them a temporary cease-fire - intended to end the fighting in this holy city . . .}\\ \hline
Class World:
\emph{ \colorunderline[color=Purple,above=0.5pt]{\text{kabul}} \colorunderline[color=Purple,above=0.5pt]{\text{,}} \colorunderline[color=Purple,above=0.5pt]{\text{afghanistan}} \colorunderline[color=Purple,above=0.5pt]{\text{-}} \colorunderline[color=Orange,above=0.5pt]{\text{government}} \colorunderline[color=Orange,above=0.5pt]{\text{troops}} \colorunderline[color=Orange,above=0.5pt]{\text{intervened}} in afghanistan ' s latest outbreak of \colorunderline[color=Red,above=0.5pt]{\text{deadly}} \colorunderline[color=Red,above=0.5pt]{\text{fighting}} \colorunderline[color=Red,above=0.5pt]{\text{between}} warlords , flying from the capital to the far west on u . s . and nato airplanes to retake an air base contested in the violence , officials said sunday . . .}\\ \hline
Class Business:
\emph{ \colorunderline[color=Cyan,above=0.5pt]{\text{london}} \colorunderline[color=Cyan,above=0.5pt]{\text{(}} \colorunderline[color=Cyan,above=0.5pt]{\text{reuters}} ) - \colorunderline[color=Magenta,above=0.5pt]{\text{the}} \colorunderline[color=Magenta,above=0.5pt]{\text{dollar}} \colorunderline[color=Magenta,above=0.5pt]{\text{dipped}} \colorunderline[color=Magenta,above=0.5pt]{\text{to}} \colorunderline[color=Magenta,above=0.5pt]{\text{a}} \colorunderline[color=Magenta,above=0.5pt]{\text{four-week}} \colorunderline[color=Magenta,above=0.5pt]{\text{low}} \colorunderline[color=Magenta,above=0.5pt]{\text{against}} \colorunderline[color=Magenta,above=0.5pt]{\text{the}} \colorunderline[color=Magenta,above=0.5pt]{\text{euro}} on monday before rising slightly on profit-taking , but steep \colorunderline[color=LimeGreen,above=0.5pt]{\text{oil}} \colorunderline[color=LimeGreen,above=0.5pt]{\text{prices}} \colorunderline[color=LimeGreen,above=0.5pt]{\text{and}} weak u . s . data continued to fan worries about the health of the world ' s largest economy .}\\ \hline
Class Business:
\emph{ new \colorunderline[color=Cyan,above=0.5pt]{\text{york}} \colorunderline[color=Cyan,above=0.5pt]{\text{(}} \colorunderline[color=Cyan,above=0.5pt]{\text{reuters}} ) - the \colorunderline[color=LimeGreen,above=0.5pt]{\text{dollar}} \colorunderline[color=LimeGreen,above=0.5pt]{\text{extended}} \colorunderline[color=LimeGreen,above=0.5pt]{\text{gains}} against the euro on monday after a report on flows into u . s . assets showed enough of a rise in \colorunderline[color=Magenta,above=0.5pt]{\text{foreign}} \colorunderline[color=Magenta,above=0.5pt]{\text{investments}} \colorunderline[color=Magenta,above=0.5pt]{\text{to}} \colorunderline[color=Magenta,above=0.5pt]{\text{offset}} \colorunderline[color=Magenta,above=0.5pt]{\text{the}} \colorunderline[color=Magenta,above=0.5pt]{\text{current}} \colorunderline[color=Magenta,above=0.5pt]{\text{account}} gap for the month .}\\ \hline
Class Business:
\emph{ reuters - \colorunderline[color=LimeGreen,above=0.5pt]{\text{apparel}} \colorunderline[color=LimeGreen,above=0.5pt]{\text{retailers}} \colorunderline[color=LimeGreen,above=0.5pt]{\text{are}} hoping theirback-to-school fashions will make the grade amongstyle-conscious teens and young adults this fall , but it couldbe a tough sell , with students and parents keeping a tighterhold on their wallets .}\\ \hline
\end{tabularx}
}
\caption{AGNews test examples correctly classified by EDUCE. Underlined set of words are excerpts extracted, one color per concept.}
\label{tab:agnewssupp1}
\end{table}

\begin{table}[t!]
\small{
\begin{tabularx}{\textwidth}{|>{\centering\arraybackslash}p{1.3cm}|X|}
  \hline
  \textcolor{Blue}{Concept 0} &
  software services giant
  /
moonwalk to home
/
video display chip
/
downloading music .
\\ \hline
 \textcolor{Gray}{Concept 1} &
upcoming my prerogative video
/
his ever-growing swimming
/
launch of a video display chip
/
illegality of downloading
\\ \hline
 \textcolor{LimeGreen}{Concept 2} &
oil market .
/
oil giant sibneft
/
oil prices and
/
ipos . public
\\ \hline
 \textcolor{Salmon}{Concept 3} &
100-meter freestyle preliminaries tuesday , a stunning blow for a
/
the red sox ' family day at
/
the winner .
/
entire era .
\\ \hline
 \textcolor{Magenta}{Concept 4} &
cash settlement of up to \#36 50 million
/
six-year deal worth about \$40 million
/
the dollar dipped to a four-week low against the euro
/
five shares ,
\\ \hline
 \textcolor{Cyan}{Concept 5} &
york , light
/
boston and texas
/
london ( reuters
/
newsday \#146 s
\\ \hline
 \textcolor{RawSienna}{Concept 6} &
olympic 100-meter freestyle
/
sox ' family
/
olympics should help
/
athletes were already
\\ \hline
 \textcolor{Red}{Concept 7} &
gunfire rattled through the
/
election . on
/
democratic coordination or cd as the
/
wiretapping internet phones to monitor criminals and terrorists is
\\ \hline
 \textcolor{Orange}{Concept 8} &
frail pope john paul
/
indian army major shot
/
unions representing workers
/
goverment representatives .
\\ \hline
 \textcolor{Purple}{Concept 9} &
lourdes , france
/
ap - american natalie
/
new york -
/
ap - it
\\ \hline
\end{tabularx}
}
\caption{Examples of excerpts that are extracted on AGNews for all $10$ concepts. Colors match the colors used in Table \ref{tab:agnewssupp1}.}
\end{table}

\begin{table}[t!]
\small{
\begin{tabularx}{\textwidth}{X}
Class Company:
\emph{ transurban manages and develops urban toll road networks in australia and north america . it \colorunderline[color=Cyan,above=0.5pt]{\text{is}} \colorunderline[color=Cyan,above=0.5pt]{ \colorunderline[color=Blue,above=0.5pt]{\text{a}}} \colorunderline[color=Cyan,above=0.5pt]{ \colorunderline[color=Salmon,above=0.5pt]{ \colorunderline[color=Blue,above=0.5pt]{\text{top}}}} \colorunderline[color=Cyan,above=0.5pt]{ \colorunderline[color=Salmon,above=0.5pt]{ \colorunderline[color=Blue,above=0.5pt]{\text{50}}}} \colorunderline[color=RawSienna,above=0.5pt]{ \colorunderline[color=Cyan,above=0.5pt]{ \colorunderline[color=Salmon,above=0.5pt]{ \colorunderline[color=Blue,above=0.5pt]{\text{company}}}}} \colorunderline[color=RawSienna,above=0.5pt]{\text{on}} \colorunderline[color=RawSienna,above=0.5pt]{\text{the}} australian securities exchange ( asx ) and has been in business since 1996 . in australia transurban has a stake in five of sydney ' s nine motorways and in melbourne it is the full owner of citylink which connects three of the city ' s major freeways . in the usa transurban has ownership interests in the 495 express lanes on a section of the capital \colorunderline[color=Magenta,above=0.5pt]{\text{beltway}} \colorunderline[color=Magenta,above=0.5pt]{\text{around}} \colorunderline[color=Magenta,above=0.5pt]{\text{washington}} \colorunderline[color=Magenta,above=0.5pt]{\text{dc}} \colorunderline[color=Magenta,above=0.5pt]{\text{.}}}\\ \hline
Class Animal:
\emph{ the red-necked falcon or red-headed merlin ( falco chicquera ) is \colorunderline[color=Red,above=0.5pt]{\text{a}} \colorunderline[color=Red,above=0.5pt]{ \colorunderline[color=RawSienna,above=0.5pt]{\text{bird}}} \colorunderline[color=Red,above=0.5pt]{ \colorunderline[color=RawSienna,above=0.5pt]{\text{of}}} \colorunderline[color=Red,above=0.5pt]{ \colorunderline[color=RawSienna,above=0.5pt]{\text{prey}}} \colorunderline[color=Red,above=0.5pt]{ \colorunderline[color=RawSienna,above=0.5pt]{\text{in}}} the falcon family . this bird is a widespread resident in india and adjacent regions as well as sub-saharan africa . it is sometimes called turumti locally . the red-necked falcon is a \colorunderline[color=Orange,above=0.5pt]{ \colorunderline[color=Salmon,above=0.5pt]{\text{medium-sized}}} \colorunderline[color=Orange,above=0.5pt]{ \colorunderline[color=Salmon,above=0.5pt]{\text{long-winged}}} \colorunderline[color=Orange,above=0.5pt]{ \colorunderline[color=Salmon,above=0.5pt]{\text{species}}} \colorunderline[color=Orange,above=0.5pt]{ \colorunderline[color=Salmon,above=0.5pt]{\text{with}}} \colorunderline[color=Orange,above=0.5pt]{ \colorunderline[color=Salmon,above=0.5pt]{\text{a}}} \colorunderline[color=Orange,above=0.5pt]{ \colorunderline[color=Salmon,above=0.5pt]{\text{bright}}} \colorunderline[color=Orange,above=0.5pt]{ \colorunderline[color=Salmon,above=0.5pt]{\text{rufous}}} \colorunderline[color=Salmon,above=0.5pt]{\text{crown}} and nape . it is on average 30–36 cm in length with a wingspan of 85 cm . the sexes are similar except in size males are smaller than females as is usual in falcons .}\\ \hline
Class Plant:
\emph{ astroloma is \colorunderline[color=Red,above=0.5pt]{\text{an}} \colorunderline[color=Red,above=0.5pt]{ \colorunderline[color=Salmon,above=0.5pt]{\text{endemic}}} \colorunderline[color=Red,above=0.5pt]{ \colorunderline[color=Salmon,above=0.5pt]{\text{australian}}} \colorunderline[color=Red,above=0.5pt]{ \colorunderline[color=Salmon,above=0.5pt]{\text{genus}}} \colorunderline[color=Red,above=0.5pt]{\text{of}} \colorunderline[color=Red,above=0.5pt]{\text{around}} 20 species of flowering \colorunderline[color=Magenta,above=0.5pt]{\text{plants}} \colorunderline[color=Magenta,above=0.5pt]{\text{in}} \colorunderline[color=Magenta,above=0.5pt]{\text{the}} family ericaceae . the majority of the species are endemic in western australia but a few species occur in new south wales victoria tasmania and south australia . species include astroloma baxteri a . cunn . ex dc . astroloma cataphractum a . j . g . wilson ms astroloma ciliatum ( lindl . ) druce astroloma compactum r . br . astroloma conostephioides ( sond . ) f . muell . ex benth .}\\ \hline
Class Album:
\emph{ stars and hank forever was the second ( and last \colorunderline[color=LimeGreen,above=0.5pt]{\text{)}} \colorunderline[color=LimeGreen,above=0.5pt]{\text{release}} \colorunderline[color=LimeGreen,above=0.5pt]{\text{in}} \colorunderline[color=Blue,above=0.5pt]{\text{the}} \colorunderline[color=Blue,above=0.5pt]{\text{american}} \colorunderline[color=Blue,above=0.5pt]{\text{composers}} series by the avant garde band the residents . the \colorunderline[color=Salmon,above=0.5pt]{\text{album}} \colorunderline[color=Salmon,above=0.5pt]{\text{was}} \colorunderline[color=Salmon,above=0.5pt]{\text{released}} in 1986 . this particular release featured a side of hank williams songs and a medley of john philip sousa marches . this was also the last studio album to feature snakefinger . kaw-liga samples the rhythm to michael jackson ' s billie jean and did well in europe it is as close as the residents ever got to a bona fide commercial hit .}\\ \hline
Class Written Work:
\emph{ fire ice is \colorunderline[color=Blue,above=0.5pt]{\text{the}} \colorunderline[color=Blue,above=0.5pt]{\text{third}} \colorunderline[color=Purple,above=0.5pt]{ \colorunderline[color=Blue,above=0.5pt]{\text{book}}} \colorunderline[color=Purple,above=0.5pt]{\text{in}} \colorunderline[color=Purple,above=0.5pt]{\text{the}} numa files series of books co-written by best-selling author clive cussler and \colorunderline[color=Salmon,above=0.5pt]{\text{paul}} \colorunderline[color=Salmon,above=0.5pt]{\text{kemprecos}} \colorunderline[color=Salmon,above=0.5pt]{\text{and}} \colorunderline[color=Salmon,above=0.5pt]{\text{was}} \colorunderline[color=Salmon,above=0.5pt]{\text{published}} in 2002 . the main character of this series is kurt austin . in this novel a russian businessman with tsarist ambitions masterminds a plot against america which involves triggering a set of earthquakes on the ocean floor creating a number of tsunami to hit the usa coastline . it is up to kurt and his team and some new allies to stop his plans .}\\ \hline
Class Educational Inst.:
\emph{ avon high school is \colorunderline[color=Blue,above=0.5pt]{\text{a}} \colorunderline[color=Orange,above=0.5pt]{ \colorunderline[color=Gray,above=0.5pt]{ \colorunderline[color=Blue,above=0.5pt]{\text{secondary}}}} \colorunderline[color=Orange,above=0.5pt]{ \colorunderline[color=RawSienna,above=0.5pt]{ \colorunderline[color=Gray,above=0.5pt]{ \colorunderline[color=Blue,above=0.5pt]{\text{school}}}}} \colorunderline[color=Orange,above=0.5pt]{ \colorunderline[color=RawSienna,above=0.5pt]{ \colorunderline[color=Gray,above=0.5pt]{\text{for}}}} \colorunderline[color=Orange,above=0.5pt]{ \colorunderline[color=RawSienna,above=0.5pt]{\text{grades}}} 9-12 located in avon \colorunderline[color=Magenta,above=0.5pt]{\text{ohio}} \colorunderline[color=Magenta,above=0.5pt]{\text{.}} \colorunderline[color=Magenta,above=0.5pt]{\text{its}} enrollment neared 1000 as of the 2008-2009 school year with a 2008 graduating class of 215 . the school colors are purple and gold . the school mascot is an eagle . the avon eagles are part of the west shore conference . they will be moving to the southwestern conference beginning in the 2015-2016 school year .}\\ \hline
Class Artist:
\emph{ vicky hamilton \colorunderline[color=LimeGreen,above=0.5pt]{\text{(}} \colorunderline[color=LimeGreen,above=0.5pt]{ \colorunderline[color=Gray,above=0.5pt]{\text{born}}} \colorunderline[color=LimeGreen,above=0.5pt]{ \colorunderline[color=Gray,above=0.5pt]{\text{april}}} \colorunderline[color=Gray,above=0.5pt]{\text{1}} \colorunderline[color=Gray,above=0.5pt]{\text{1958}} ) is \colorunderline[color=Blue,above=0.5pt]{\text{an}} \colorunderline[color=Blue,above=0.5pt]{\text{american}} \colorunderline[color=Blue,above=0.5pt]{\text{record}} \colorunderline[color=Blue,above=0.5pt]{\text{executive}} \colorunderline[color=Blue,above=0.5pt]{\text{personal}} \colorunderline[color=Blue,above=0.5pt]{\text{manager}} \colorunderline[color=Blue,above=0.5pt]{\text{promoter}} and club booker writer ( journalist playwright and screenwriter ) documentary film maker and artist . hamilton is noted for managing the early careers of guns n ' roses poison and faster pussycat for being a management consultant for mötley crüe and stryper a 1980s concert promoter on the sunset strip and a club booker at bar sinister from 2001 to 2010 . [...]} \\ \hline
Class Athlete:
\emph{ james jim arthur bacon ( \colorunderline[color=Gray,above=0.5pt]{\text{birth}} \colorunderline[color=Gray,above=0.5pt]{\text{registered}} \colorunderline[color=Gray,above=0.5pt]{\text{october–december}} \colorunderline[color=Gray,above=0.5pt]{\text{1896}} in newport district — death unknown ) was \colorunderline[color=Red,above=0.5pt]{\text{a}} \colorunderline[color=Red,above=0.5pt]{\text{welsh}} \colorunderline[color=Red,above=0.5pt]{\text{rugby}} \colorunderline[color=Red,above=0.5pt]{\text{union}} \colorunderline[color=Red,above=0.5pt]{\text{and}} \colorunderline[color=Red,above=0.5pt]{\text{professional}} \colorunderline[color=Red,above=0.5pt]{\text{rugby}} \colorunderline[color=Red,above=0.5pt]{\text{league}} \colorunderline[color=Red,above=0.5pt]{\text{footballer}} \colorunderline[color=Red,above=0.5pt]{\text{of}} the 1910s and ' 20s and coach of the 1920s playing club level rugby union ( ru ) for cross keys rfc and representative level rugby league ( rl ) for great britain and wales and at club level for leeds as a wing or centre i . e . number 2 or 5 or 3 or 4 and coaching club level rugby league ( rl ) for castleford .}\\ \hline
Class Office Holder:
\emph{ jonathan david morris \colorunderline[color=LimeGreen,above=0.5pt]{\text{(}} \colorunderline[color=LimeGreen,above=0.5pt]{ \colorunderline[color=Gray,above=0.5pt]{\text{october}}} \colorunderline[color=LimeGreen,above=0.5pt]{ \colorunderline[color=Gray,above=0.5pt]{\text{8}}} \colorunderline[color=LimeGreen,above=0.5pt]{ \colorunderline[color=Gray,above=0.5pt]{\text{1804}}} \colorunderline[color=LimeGreen,above=0.5pt]{ \colorunderline[color=Gray,above=0.5pt]{\text{-}}} \colorunderline[color=LimeGreen,above=0.5pt]{ \colorunderline[color=Gray,above=0.5pt]{\text{may}}} \colorunderline[color=LimeGreen,above=0.5pt]{ \colorunderline[color=Gray,above=0.5pt]{\text{16}}} \colorunderline[color=Gray,above=0.5pt]{\text{1875}} ) was a u . s . representative from ohio son of thomas morris and brother of isaac n . morris . born in columbia hamilton county ohio morris attended the public schools . he studied law . he was admitted to the bar and commenced practice in batavia ohio . he served as clerk of the courts of clermont county . morris was elected as a \colorunderline[color=Magenta,above=0.5pt]{\text{democrat}} \colorunderline[color=Magenta,above=0.5pt]{\text{to}} \colorunderline[color=Magenta,above=0.5pt]{\text{the}} thirtieth congress to fill the vacancy caused by the death of thomas l .}\\ \hline
Class Mean Of Transp.:
\emph{ german submarine u-32 \colorunderline[color=Cyan,above=0.5pt]{\text{was}} \colorunderline[color=Red,above=0.5pt]{ \colorunderline[color=Cyan,above=0.5pt]{ \colorunderline[color=Blue,above=0.5pt]{\text{a}}}} \colorunderline[color=Red,above=0.5pt]{ \colorunderline[color=Cyan,above=0.5pt]{ \colorunderline[color=Blue,above=0.5pt]{\text{type}}}} \colorunderline[color=Red,above=0.5pt]{ \colorunderline[color=Cyan,above=0.5pt]{ \colorunderline[color=Blue,above=0.5pt]{\text{viia}}}} \colorunderline[color=Purple,above=0.5pt]{ \colorunderline[color=Red,above=0.5pt]{ \colorunderline[color=Cyan,above=0.5pt]{ \colorunderline[color=Blue,above=0.5pt]{\text{u-boat}}}}} \colorunderline[color=Purple,above=0.5pt]{ \colorunderline[color=Red,above=0.5pt]{\text{of}}} \colorunderline[color=Purple,above=0.5pt]{\text{nazi}} germany ' s kriegsmarine during world war ii . her keel was laid down on 15 march 1936 by ag weser of bremen as werk 913 . she was launched on 25 february 1937 and commissioned on 15 april with kapitänleutnant ( kptlt . ) werner lott in command . on 15 august 1937 lott was relieved by korvettenkapitän ( krv . kpt . ) paul büchel and on 12 february 1940 oberleutnant zur see ( oblt . z . s . ) hans jenisch took over he was in charge of the boat until her loss .}\\ \hline
Class Building:
\emph{ the tootell house ( also called king ' s row or hedgerow ) is \colorunderline[color=Blue,above=0.5pt]{\text{a}} \colorunderline[color=Blue,above=0.5pt]{\text{house}} \colorunderline[color=Blue,above=0.5pt]{\text{at}} 1747 \colorunderline[color=Magenta,above=0.5pt]{\text{mooresfield}} \colorunderline[color=Magenta,above=0.5pt]{\text{road}} \colorunderline[color=Magenta,above=0.5pt]{\text{in}} \colorunderline[color=Magenta,above=0.5pt]{\text{kingston}} \colorunderline[color=Magenta,above=0.5pt]{\text{rhode}} \colorunderline[color=Magenta,above=0.5pt]{\text{island}} \colorunderline[color=Magenta,above=0.5pt]{\text{that}} is listed on the \colorunderline[color=Orange,above=0.5pt]{\text{national}} \colorunderline[color=Orange,above=0.5pt]{\text{register}} \colorunderline[color=Orange,above=0.5pt]{ \colorunderline[color=Cyan,above=0.5pt]{ \colorunderline[color=LimeGreen,above=0.5pt]{\text{of}}}} \colorunderline[color=Orange,above=0.5pt]{ \colorunderline[color=Cyan,above=0.5pt]{ \colorunderline[color=LimeGreen,above=0.5pt]{\text{historic}}}} \colorunderline[color=Orange,above=0.5pt]{ \colorunderline[color=Cyan,above=0.5pt]{ \colorunderline[color=LimeGreen,above=0.5pt]{\text{places}}}} \colorunderline[color=Cyan,above=0.5pt]{ \colorunderline[color=LimeGreen,above=0.5pt]{\text{.}}} \colorunderline[color=Cyan,above=0.5pt]{\text{the}} \colorunderline[color=Cyan,above=0.5pt]{\text{two-story}} \colorunderline[color=Cyan,above=0.5pt]{\text{wood-shingled}} colonial revival house on a 3-acre ( 12000 m2 ) tract was designed by gunther and beamis associates of boston for mr . \& mrs . f . delmont tootell and was built in 1932-1933 . house design was by john j . g . gunther and elizabeth clark gunther was the landscape architect for the grounds .}\\ \hline
Class Village:
\emph{ angamoozhi \colorunderline[color=Cyan,above=0.5pt]{\text{is}} \colorunderline[color=Cyan,above=0.5pt]{\text{a}} \colorunderline[color=Orange,above=0.5pt]{ \colorunderline[color=Cyan,above=0.5pt]{ \colorunderline[color=Gray,above=0.5pt]{\text{village}}}} \colorunderline[color=Orange,above=0.5pt]{ \colorunderline[color=Gray,above=0.5pt]{\text{in}}} \colorunderline[color=Orange,above=0.5pt]{ \colorunderline[color=Gray,above=0.5pt]{\text{pathanamthitta}}} \colorunderline[color=Orange,above=0.5pt]{\text{district}} located in \colorunderline[color=Magenta,above=0.5pt]{\text{kerala}} \colorunderline[color=Magenta,above=0.5pt]{\text{state}} \colorunderline[color=Magenta,above=0.5pt]{\text{india}} \colorunderline[color=Magenta,above=0.5pt]{\text{.}} angamoozhi is near seethathodu town . geographically angamoozhi is a high-range area . it is mainly a plantation township . both state run ksrtc and private operated buses connect angamoozhi to pathanamthitta city . tourist can avail the travelling facility by ksrtc service ( morning 5 30 from kumili and 11 30 from pathanamthitta ) in between kumili and pathanamthitta via vallakkadavu angamoozhi kakki dam and vadaserikkara and can enjoy the beauty of the forest . [ citation needed ]}\\ \hline
\end{tabularx}
}
\caption{DBPedia test examples correctly classified by EDUCE. Underlined set of words are excerpts extracted, one color per concept.}
\label{tab:dbsupp1}
\end{table}

\begin{table}[t!]
\small{
\begin{tabularx}{\textwidth}{|>{\centering\arraybackslash}p{1.3cm}|X|}
  \hline
  \textcolor{Blue}{Concept 0} &
  a top 50 company
  /
  the regular seasonal movement
  /
  the production of port wine
  /
  the american composers
  /
  a bollywood film
  /
  the third book
  /
  a secondary school
  /
  an american record executive personal manager promoter
  /
  a well-known v8 supercars presenter and commentator
  /
  a merchant and public official
  /
  a type viia u-boat
  /
  a house at
  /
  the main fort
  /
  a major poet
  \\ \hline
   \textcolor{Gray}{Concept 1} &
  born in philadelphia
  /
  tranquil star ( foaled
  /
  instructional documentary produced
  /
  bollywood film released in
  /
  nineteenth instalment in
  /
  secondary school for
  /
  born april 1 1958
  /
  birth registered october–december 1896
  /
  october 8 1804 - may 16 1875
  /
  academy is the volleyball academy of
  /
  village in pathanamthitta
  \\ \hline
   \textcolor{LimeGreen}{Concept 2} &
  ( 1800–1870 ) probably born in
  /
  ( 12 april
  /
  ) release in
  /
  is a bollywood film released in
  /
  an omnibus release from
  /
  a studio school located in
  /
  ( born april
  /
  gunji 8 july
  /
  ( october 8 1804 - may 16
  /
  an iron collier in
  /
  of historic places .
  /
  fort ( also called
  \\ \hline
   \textcolor{Salmon}{Concept 3} &
  top 50 company
  /
  medium-sized long-winged species with a bright rufous crown
  /
  endemic australian genus
  /
  album was released
  /
  bollywood film released
  /
  paul kemprecos and was published
  /
  inns of court exclusively entitled
  /
  debut album explores
  /
  lagarto ( lizard
  /
  hong kong cross-media creator
  /
  bird park in
  /
  wild blackwater river
  \\ \hline
   \textcolor{Magenta}{Concept 4} &
  beltway around washington dc .
  /
  democratic republic of
  /
  plants in the
  /
  party live in
  /
  my texas .
  /
  asheville north carolina .
  /
  ohio . its
  /
  executive . he
  /
  democrat to the
  /
  south africa .
  /
  mooresfield road in kingston rhode island that
  /
  north to south .
  /
  kerala state india .
  \\ \hline
   \textcolor{Cyan}{Concept 5} &
  is a top 50 company
  /
  is a genus of crane fly
  /
  is a red portuguese wine grape
  /
  is the debut studio record
  /
  is an abc family original movie
  /
  was a regional fashion entertainment and lifestyle publication
  /
  interactive technologies institute
  /
  is a beirut-based lebanese haute couture fashion designer
  /
  was a montenegrin serb boxer
  /
  is the chief technology officer
  /
  was a type viia u-boat
  /
  of historic places . the two-story wood-shingled
  /
  fort ( also
  /
  is a village
  \\ \hline
   \textcolor{RawSienna}{Concept 6} &
  company on the
  /
  bird of prey in
  /
  apple . in
  /
  documentary produced by
  /
  film released in
  /
  sequel to the
  /
  school for grades
  /
  animator . he
  /
  catcher . he
  /
  cross-media creator who
  /
  aircraft corporation to
  /
  museum was made in
  \\ \hline
   \textcolor{Red}{Concept 7} &
  the brands club mahindra holidays club mahindra travel club
  /
  a bird of prey in
  /
  an endemic australian genus of around
  /
  snooker world rankings 1978/1979 the professional world rankings for
  /
  a black and white chinese animation short made in
  /
  a british rugby league periodical that
  /
  a russian youth football academy based in
  /
  a turkish cypriot singer and athlete .
  /
  a welsh rugby union and professional rugby league footballer of
  /
  a planter a confederate cavalry general in
  /
  a type viia u-boat of
  /
  a volleyball academy .
  /
  an uninhabited steep rocky island west of
  \\ \hline
   \textcolor{Orange}{Concept 8} &
  street sounds was
  /
  medium-sized long-winged species with a bright rufous
  /
  rataj places it
  /
  ep by linkin park
  /
  white chinese animation
  /
  british schools (
  /
  secondary school for grades
  /
  argentine painter of the concretist and cubist schools
  /
  eminent anglican priest
  /
  hospital ships was built
  /
  national register of historic places
  /
  uninhabited steep rocky island west
  /
  village in pathanamthitta district
  \\ \hline
   \textcolor{Purple}{Concept 9} &
  translation pension fund for
  /
  adaptation which enables
  /
  mopeds produced by
  /
  instructional documentary produced by
  /
  bollywood film released in
  /
  book in the
  /
  river in kwazulu-natal
  /
  memoir of her
  /
  boxer in the
  /
  presidency of dmitry medvedev began
  /
  u-boat of nazi
  /
  harbour of helsingør dedicated to
  /
  uninhabited steep rocky island west of
  \\ \hline
\end{tabularx}
}
\caption{Examples of excerpts that are extracted on DBPedia for all $10$ concepts. Colors match the colors used in Table \ref{tab:dbsupp1}.}
\end{table}

\clearpage
\section{Applying EDUCE principles for Image Classification}
\label{imludo}
Having assessed the relevance of our model on text data, we now turn to image data and explore if EDUCE is also able to extract meaningful concepts. In that case, the excerpts extracted in textual data are replaced by regions (or patches) in an image, and the recurrent neural architecture processing text is replaced by a classical convolutional neural network.

\paragraph{Principles: } Each image $x$ is first encoded through a convolutional neural network that output a representation in $h(x)=\mathbb{R}^{W \times H \times D}$ where  $D$ is the number of output filters. Each vector $h_{i,j,.}(x) \in \mathbb{R}^D$ corresponds to a particular region $s$ of the input image (i.e the receptive field of the convolutional neural network). Then each $h_{i,j,.}$ is used to compute $p_\gamma(s|x,c)$ through a linear model and a softmax activation as in Section 2.1 (step 1), and to compute $p_\alpha(z_c=1|s_c(x),c)$ (Section 2.1, step 2) here again with a linear model and a Bernoulli distribution.  The semantics homogeneity is ensured by using an image classification network over the region captured by the receptive field in the previous convolutional neural networks that ensure that all regions captured from a particular concept ''look the same'' and look differently from the regions captured for other concepts.

\begin{figure}[t!]
\begin{center}
\includegraphics[width=0.9\linewidth]{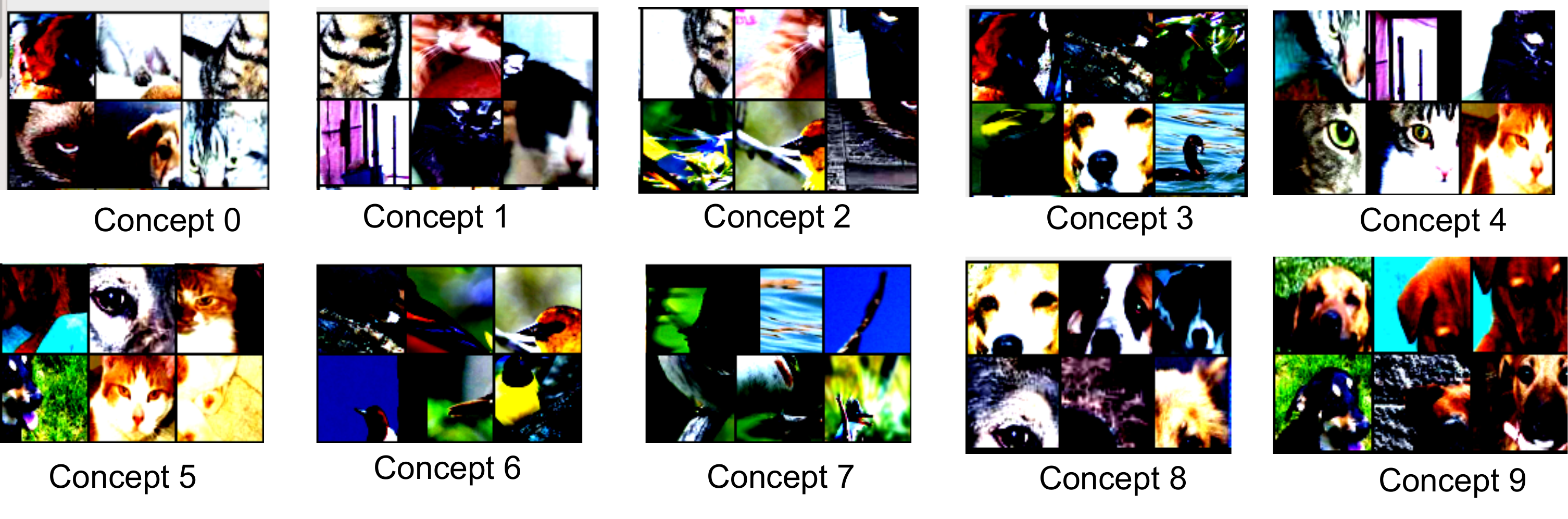}
\end{center}
\caption{Samples of patterns extracted for each concept in the birds-cats-dogs dataset.}
\label{fig:m2}
\end{figure}

\begin{figure}[t!]
\begin{subfigure}[b]{0.45\textwidth}
        \centering \includegraphics[width=\textwidth]{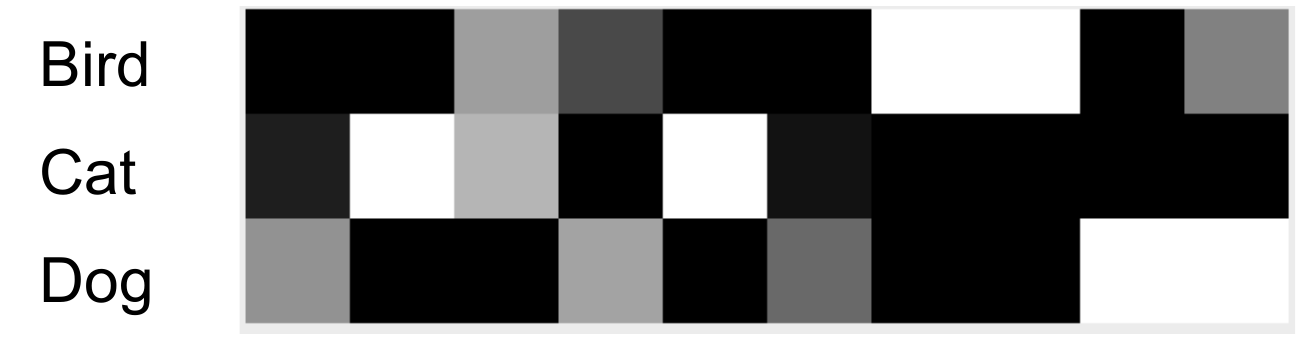}
    \caption{Final Classifier weights, each column is a concept (0 to 9, from left to right) each row a final category. The lighter, the higher the weight is. Concepts 6 and 7 are associated to category 'Bird', Concept 1 and 4 to 'Cat' and concepts 8 and 9 to 'Dog'. }
  \label{fig:m3}
\end{subfigure}
\hfill
\begin{subfigure}[b]{0.45\textwidth}
        \centering \includegraphics[width=\textwidth]{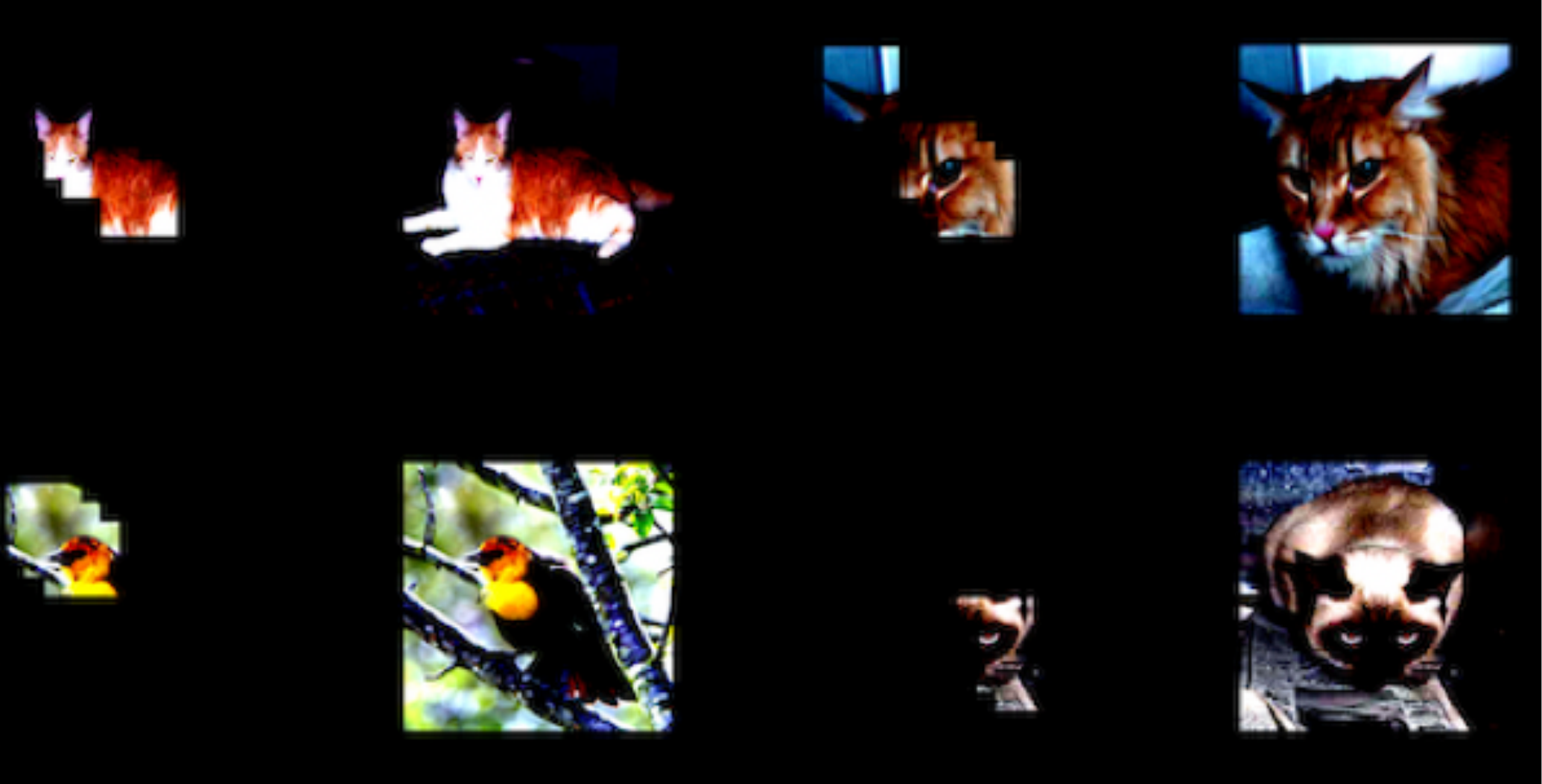}
        \caption{Examples of classified images (column 2 and 4) and corresponding extracted patterns (column 1 and 3) for concept detected as present.}
          \label{fig:m4}
    \end{subfigure}
\caption{Interpretation of EDUCE on images.}
\end{figure}

\paragraph{Experiments: }
To ensure the validity of our adaptation of EDUCE to images, we conduct very simple experiments on a dataset composed of $224\times224$ RGB images split in $3$ categories: dogs, cats, and birds\footnote{We construct this dataset by combining random images from the Caltech Bird 200-2011 dataset \citep{WahCUB_200_2011} with images of the cats-and-dogs Kaggle dataset \citep{KaggleDogVCat}} in equal proportion. We train on $3,000$ images and test on $3,000$ images. We build our model on top of a pretrained VGG-11 model \cite{Simonyan14c} which is used for computing the intermediate representations $h(x)$. Said otherwise, the output of the VGG-11 model is used for both detecting relevant regions and thus building the binary representation $z$, but also to ensure the homogeneity of extracted regions during training. Figure \ref{fig:m2} shows extracted patterns and associated categories. Final classification performance is $91.6\%$ with $10$ concepts. In Figure \ref{fig:m2} we plot extracted patterns for the $10$ concepts and report in Figure \ref{fig:m3} the weights of the final classifier. From these two figures, we can interpret the model's behavior: concept 8 and concept 9 show what differentiates a \emph{dog} from a \emph{cat} or a \emph{bird}, and support the classifier's prediction of the dog category. Moreover, by looking at which regions are extracted by our model (Figure \ref{fig:m4} ), we can see that EDUCE can focus its attention to discriminative regions of the images.

\end{document}